%% file: main.tex
\documentclass[runningheads]{llncs}
\input{preamble}

\begin{document}
\title{GOMP-ST: Grasp Optimized Motion Planning for Suction Transport}
\author{Yahav Avigal*\inst{}
\and
Jeffrey Ichnowski*\inst{}
\and
Max Yiye Cao\inst{}
\and
Ken Goldberg\inst{}\\
\scriptsize{* Equal contribution}}
\authorrunning{Avigal et al.}
\institute{University of California, Berkeley, USA}
\maketitle              %
\input{abstract}
\input{section-1-introduction}

\input{section-2-related_work}
\input{section-3-problem_statement}
\input{section-4-method}
\input{section-5-experiments}
\input{section-6-conclusion}
\input{section-7-acks}

\bibliographystyle{splncs04}
{ %
\renewcommand{\clearpage}{} 
\vspace{48pt}
\bibliography{references}
}

\end{document}

%% file: preamble.tex
\usepackage{times}

\usepackage[T1]{fontenc}
\def\doi#1{\href{https://doi.org/\detokenize{#1}}{\url{https://doi.org/\detokenize{#1}}}}
\usepackage{graphicx}
\usepackage{listings}
\usepackage[numbers]{natbib}
\usepackage{multicol}
\usepackage{multirow}
\usepackage[bookmarks=true]{hyperref}
\usepackage{amsmath}
\usepackage{amsfonts}
\usepackage{graphicx}
\usepackage{multirow}
\usepackage{booktabs}
\usepackage{textcomp}
\usepackage[dvipsnames]{xcolor}
\usepackage{tikz}
\usetikzlibrary{positioning,shapes.geometric,arrows,fit,shapes.symbols,svg.path,tikzmark,calc}

\pgfdeclarelayer{background}
\pgfdeclarelayer{foreground}
\pgfsetlayers{background,main,foreground}

\usepackage{caption}
\usepackage{subcaption}
\newcommand{\algfull}{Grasp Optimized Motion Planning for Suction Transport}
\newcommand{\algname}{GOMP-ST}

\renewcommand{\vec}[1]{\mathbf{#1}}

%% file: abstract.tex
\begin{abstract}
Suction cup grasping is very common in industry, but moving too quickly can cause suction cups to detach, causing drops or damage. Maintaining a suction grasp throughout a high-speed motion requires balancing suction forces against inertial forces while the suction cups deform under strain.  In this paper, we consider Grasp Optimized Motion Planning for Suction Transport (GOMP-ST), an algorithm that combines deep learning with optimization to decrease transport time while avoiding suction cup failure. GOMP-ST first repeatedly moves a physical robot, vacuum gripper, and a sample object, while measuring pressure with a solid-state sensor to learn critical failure conditions.  Then, these are integrated as constraints on the accelerations at the end-effector into a time-optimizing motion planner. The resulting plans incorporate real-world effects such as suction cup deformation that are difficult to model analytically. %
In GOMP-ST, the learned constraint, modeled with a neural network, is linearized using Autograd and integrated into a sequential quadratic program optimization. In 420 experiments with a physical UR5 transporting objects ranging from 1.3 to 1.7\,kg, we compare GOMP-ST to baseline optimizing motion planners.  Results suggest that GOMP-ST can avoid suction cup failure while decreasing transport times from 16\,\% to 58\,\%. For code, video, and datasets, see
\url{https://sites.google.com/view/gomp-st} 

\keywords{Motion Planning \and Manipulation \and Optimization.}
\end{abstract}

%% file: section-1-introduction.tex
\section{Introduction}
Vacuum suction cup grasping, due to its ability to quickly hold and release a large variety of objects, is a common grasping modality for robots in industrial settings such as warehouses and logistics centers.  With the recent rise in demand for robot pick-and-place operations, the speed of object transport is critical.  However, suction grasps can fail if the object is transported too quickly.  Determining the conditions where suction grasps fail is non-trivial due to the difficult-to-model deformations of suction cups under stress. Existing analytic models make simplifying assumptions, such as rigid suction cups~\cite{pham2013kinodynamic} or quasi-static physics~\cite{mahler2018dex, pham2019critically, huh2021multi}. An alternative is to heuristically slow motions when objects are held in suction grasps.

In prior work, the Grasp-Optimized Motion Planner (GOMP~\cite{ichnowski2020gomp}) leveraged an unconstrained degree of freedom (DoF) around the grasp axis to optimize pick-and-place motions for parallel jaw grippers.
Grasp-Optimized Motion Planning for Fast Inertial Transport (GOMP-FIT~\cite{ichnowski2022gompfit}) computes time-optimized motions while taking into account end-effector and object acceleration constraints to reduce product damage and spills. However, when executed on robots with suction grippers, the resulting motions can fail due to suction cups detaching.

\input{figures/figure_one}

To address this problem, we propose the \algfull{} (\algname{}), an algorithm that computes time-optimized motions by integrating a \emph{learned} suction grasp loss constraint.
\algname{} first tries varied rapid lifts of a given object with suction grasps to find motions that cause suction failures. It then learns a model from the data based on a history, or sequence, of end-effector accelerations to define a constraint function.
At run time, the optimization treats the learned model as a non-linear constraint on the motion by linearizing it and computing a first-order approximation based on its Jacobian.  In our implementation, we use a neural network to learn the suction model and use Autograd to obtain the Jacobian.

In experiments with steel rectangular blocks, we learn a model of suction failure on a physical UR5 with a 4-cup vacuum gripper.  We then apply the learned model to transport 4 objects of varying mass held in suction between multiple different start and goal pairs and around obstacles.  We compare \algname{} to GOMP, GOMP-FIT with an analytic model of suction, and ablations of \algname{}.  We find that \algname{} can achieve a near 100\,\% success rate, for motions that are 16 to 58\,\% faster. 

The contributions of this paper are:
\begin{itemize}
    \item A novel algorithm, \algname{}, \algfull{}, based on:
    \begin{enumerate}
    \item Formulation of a learnable acceleration constraint for suction cup transport
    \item An efficient method for learning the constraint via boundary searching and data augmentation
    \item Integration of the learned constraint into an optimizing motion planner
    \end{enumerate}
    \item Data from experiments with a physical robot comparing \algname{} to baselines.
\end{itemize}

%% file: figures/figure_one.tex
\begin{figure}[t]
    \centering
    \begin{tabular}{cc}
        \includegraphics[height=130pt,clip,trim=90 0 290 0]{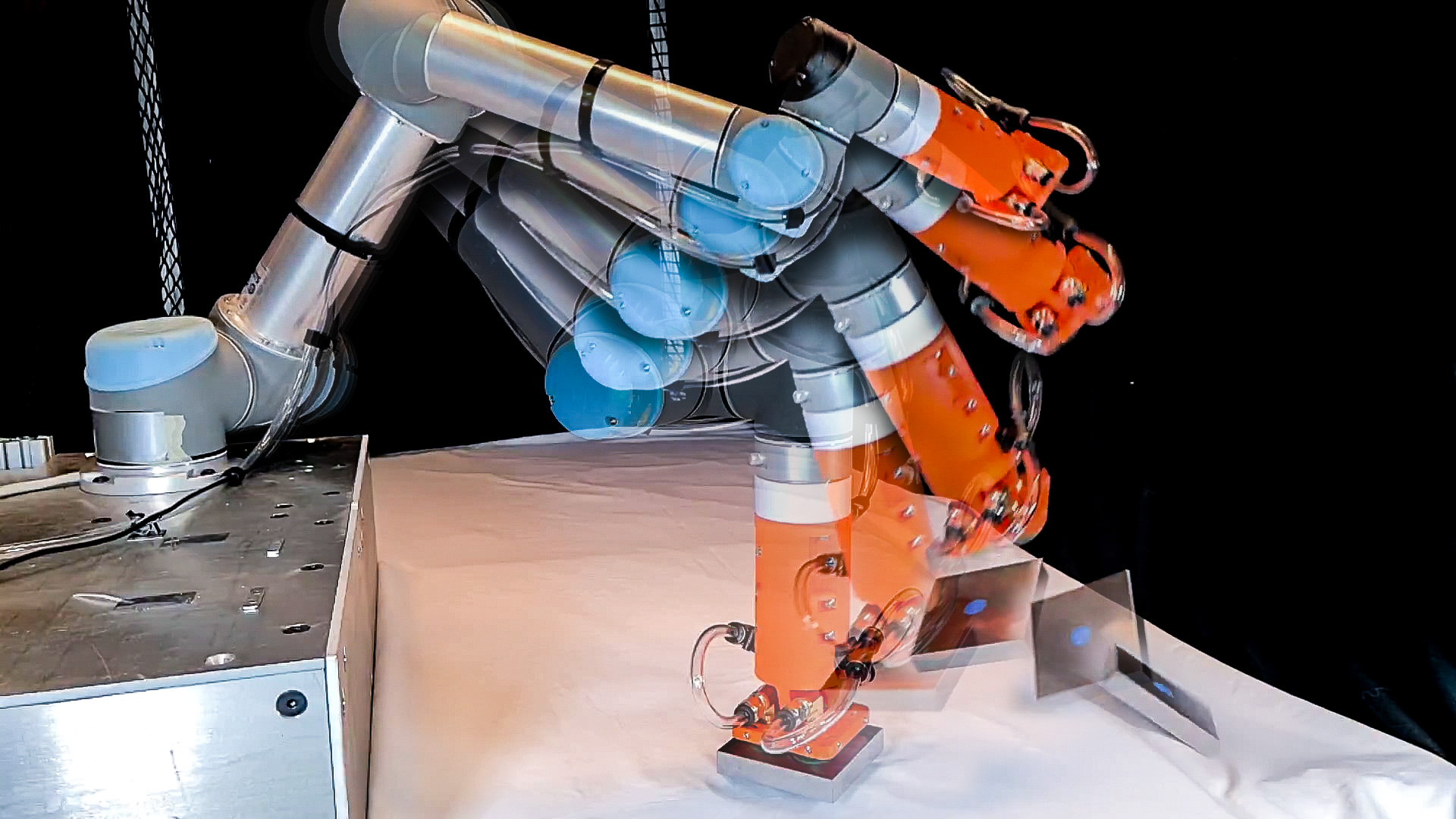} &
        \includegraphics[height=130pt,clip,trim=600 70 200 70]{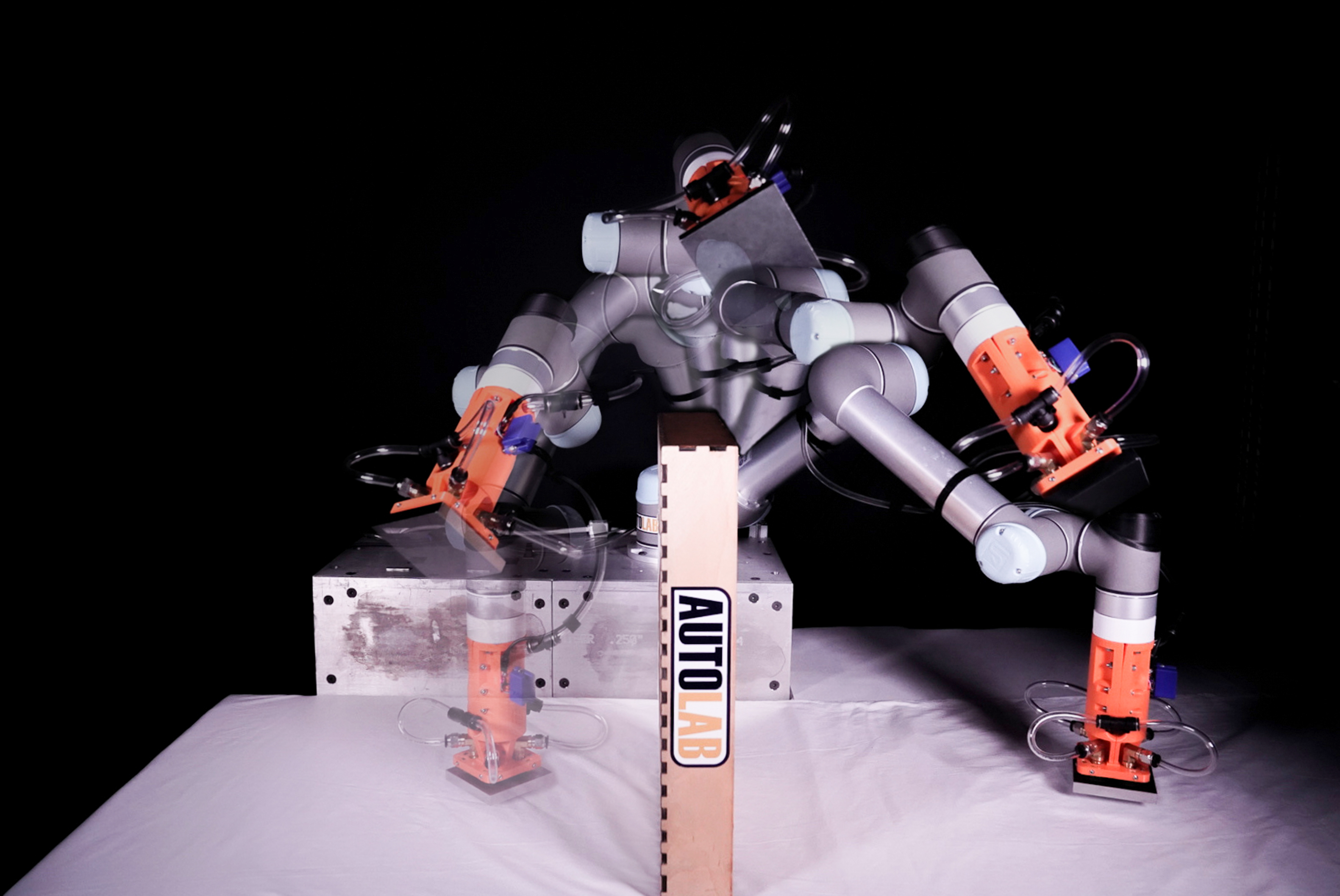} \\
        (a) A fast motion resulting in suction failure & (b) A successful transport over obstacle
    \end{tabular}
    \caption{\textbf{High-speed motions can cause failure of suction cups.}
    These multiple-exposure images show high-speed motions computed by the time-optimizing motion planners (a) J-GOMP and (b) \algname{}. The fast trajectory computed by J-GOMP and inertial forces cause the suction grasp to fail and the grasped block to fall away and to the right.
    In this paper, we propose \algname{}, an algorithm that incorporates a learned acceleration constraint into a time-optimizing motion planner to avoid such failures.}
    \label{fig:figure_one}
\end{figure}

%% file: section-2-related_work.tex
\section{Related Work}

\subsection{Motion Planning and Optimization}
Robot motion planning aims to find safe and efficient robot motions from a start to a goal configuration that avoid obstacles. Sampling-based motion planners, such as PRM~\cite{Kavraki1996_TRA} and bi-directional RRT~\cite{LaValle2000_WAFR} have variants that are probabilistically-complete and asymptotically-optimal. %
Optimization-based motion planners, such as TrajOpt~\cite{schulman2013finding}, STOMP~\cite{kalakrishnan2011stomp},  CHOMP~\cite{ratliff2009chomp}, KOMO~\cite{toussaint2014newton}, and ITOMP~\cite{park2012itomp} can compute optimized trajectories iteratively improving paths or interleaving with sampling-based planners~\cite{kuntz2020fast}.  Grasp-Optimized Motion Planning (GOMP)~\cite{ichnowski2020gomp} and Deep-Jerk GOMP (DJ-GOMP)~\cite{ichnowski2020djgomp} leverage a degree-of-freedom in the grasp pose while integrating dynamic and kinematic constraints in a sequential quadratic program that iteratively computes time-optimized pick-and-place motions. GOMP for Fast Inertial Transport (GOMP-FIT)~\cite{ichnowski2022gompfit} incorporates end-effector acceleration constraints for parallel-jaw grippers by employing the forward pass of the Recursive Newton Euler (RNE) method~\cite{luh1980line}. In contrast to prior work, \algname{} learns a suction-cup constraint to avoid suction grasp failure.

\subsection{Constraint Learning in Optimization}
Optimization using constraints based on empirical models is gaining traction. 
Maragno et al.~\cite{maragno2021mixed} integrate learned constraints into a mixed-integer optimization using trust regions defined by the convex hull of the training data.
In contrast, we propose using domain knowledge to perform data augmentation in a continuous optimization.
Kud{\l}a et al.~\cite{kudla2018one} propose a learning a decisions tree of constraints and a transform appropriate for mixed integer linear programming.  While the decision tree provides flexible conditions for constraints, we instead propose exploiting the nature of the trajectory optimization by regressing on inputs containing the last $h$ time steps.
Bartolini et al.~\cite{bartolini2011neuron} and later Lombardi et al.~\cite{lombardi2017empirical} show that neural networks can learn constraints based on difficult-to-model phenomena be integrated into constrained combinatorial optimizations.  We employ neural network constraints in continuous non-convex optimization. %
De Raedt et al.~\cite{deRaedt2018learning} provide a recent survey of constraint acquisition, and 
Fajemisin et al.~\cite{fajemisin2021optimization} provide a recent survey of optimization with constraint learning.  These surveys provide a wealth of ideas that could be extended to apply additional constraint learning and learned constraints to optimizing motion planning.

\subsection{Suction Grasping}
Suction grasping is widely used in industrial settings. Due to the increasing interest in suction grasping for pick-and-place tasks, recent models focus on computing a robust suction grasp. Dex-Net 3.0~\cite{mahler2018dex} introduced a novel suction contact model that quantifies seal formation using a quasi-static spring system, along with a robust version of the model under random disturbing wrenches and perturbations in object pose, gripper pose, and friction. Huh et al.~\cite{huh2021multi} use a learned model and a novel multi-chamber suction cup design to detect failures in the suction seal early to avoid grasp failures. However, these works assume quasi-static physics, an assumption that does not hold for high inertial forces.

Most closely related to this paper is the work by Pham and Pham~\cite{pham2019critically} which proposes a suction-cup model and identify a contact stability constraint and a pipeline to parameterize time-optimal geometric paths satisfying the constraint. %
\algname{} differs in a few ways.  First, their method first plans a motion using a sampling-based planner, then time-parameterizes the motion, whereas \algname{} integrates planning and time-parameterization into a single optimization.  Second, their method uses an analytic model that does not include suction cup deformation, whereas \algname{} learns a constraint based on suction grasps failures through experimentation.  Finally, since their method computes path and timing in separate steps, it does not explore alternate paths with potentially better timing; whereas \algname{} integrates planning, time-optimization, and suction constraints simultaneously to adjust the path and timing together.

\subsection{Dynamic Manipulation}

Dynamic manipulation exploits forces due to accelerations, along with kinematics, static,
and quasi-static forces to achieve a task~\cite{ruggiero2018nonprehensile}. Lynch and Mason~\cite{lynch1996dynamic} leverage centrifugal and Coriolis forces to allow low degree-of-freedom robots to control objects with more degrees-of-freedom.
Lynch and Mason~\cite{lynch1999dynamic} also directly integrate constraints in a sequential quadratic problem to plan robot trajectories that achieve a dynamic task, such as snatching an object, throwing, and rolling, via coupling forces through the non-prehensile contact in an obstacle-free environment.
Srinivasa et al.~\cite{srinivasa2005using} address the problem of rolling a block
resting on a flat palm by employing constraints on accelerations at the contact point.
Mucchiani and Yim~\cite{mucchiani2021dynamic} propose a grasping approach that utilizes object inertia for sweeping an object at rest to a goal position by leveraging accelerations and torques from the path as stabilizing forces in the passive end-effector.
In this work, we employ constraints on the end-effector accelerations to perform efficient dynamic object transport around obstacles using a suction gripper.

One promising related avenue of research investigates motion planning and optimization that integrates forces present at the end-effector.  
Hauser~\cite{hauser2014fast} investigates including contact forces during the optimization, and Luo and Hauser~\cite{luo2017robust} extend this to include a learned confidence into an optimization.  In a similar manner, Bernheisel and Lynch~\cite{bernheisel2004stable} and Acharya et al.~\cite{acharya2020nonprehensile} both explore different ways to address the \emph{waiter's problem} which requires generating motions of a tray to perform non-prehensile balancing of objects.  In contrast to these lines of work, we focus on learning and integrating a constraint to maintain suction contact as a differentiable function, instead of learning the parameters of a model and integrating it into a plan.

When the dynamics are unknown or the existing models are insufficiently accurate, a promising approach is to leverage data-driven methods.
Zeng et al.~\cite{zeng2020tossingbot} propose TossingBot that learns parameters of a pre-defined dynamic motion to toss objects into target bins using parallel-jaw end-effectors.
Wang et al.~\cite{wang2020swingbot} propose SwingBot, that uses tactile feedback to learn how to dynamically swing up novel objects.
In Robots of the Lost Arc~\cite{zhang2021rotla}, a robot computes high-speed motions to induce fixed-end cables to swing over distant obstacles, knock down target objects, and weave between obstacles.
Ha and Song~\cite{ha2021flingbot} propose FlingBot, a self-supervised learning framework to learn dynamic flinging actions for cloth unfolding. 
Lim et al.~\cite{lim2021prc}, uses simulation and physical data to train a model for swinging an object to hit a target.
In this work, we integrate a learned a constraint in the solver allowing it to compute fast motions while maintaining the suction seal.

%% file: section-3-problem_statement.tex
\section{Problem Statement}

Let $\vec{q} \in \mathcal{C}$ be the complete specification of the degrees of freedom for a robot, where $\mathcal{C}$ is the set of all possible configurations. 
Let $\mathcal{O}$ be the set of obstacles, and $\mathcal{C}_\mathrm{obs} \subseteq \mathcal{C}$ be the set of configurations in collision with $\mathcal{O}$. Let $\mathcal{C}_\mathrm{free} = \mathcal{C} \setminus \mathcal{C}_\mathrm{obs}$ be the set of configurations that are not in collision. 
Let $\dot{\vec{q}}$ and $\ddot{\vec{q}}$ be the first and second derivatives of the configuration. 
Let $\vec{a}_\mathrm{ee} \in \mathbb{R}^3$ be the linear acceleration of the end effector with one or more suction cups holding a known object $b$, and $\hat{\vec{n}} \in \mathbb{R}^3$ be the suction cup normal, as shown in Fig.~\ref{fig:fbd}

Given a start $\vec{q}_0$ and goal $\vec{q}_\mathrm{goal}$ configuration, the objective of \algname{} is to compute a trajectory $\tau = (\vec{x}_0, \vec{x}_1, \ldots, \vec{x}_{H})$, where $\vec{x}_t = (\vec{q}_t, \dot{\vec{q}}_t, \ddot{\vec{q}}_t) \in \mathcal{X}$ is the state of the robot at time $t$, $\mathcal{X}$ is the set of states, such that $\vec{q}_t \in \mathcal{C}_\mathrm{free}\; \forall t \in [0, H]$, and $\dot{\vec{q}}$ and $\ddot{\vec{q}}$ are within the box-bounded dynamic limits of the robot, and $\vec{q}_H  = \vec{q}_\mathrm{goal}$.  The object $b$ will remain attached throughout the motion.  As with GOMP, start and goal configurations may also be expressed via forward kinematics and bounds on degrees of freedom.

%% file: section-4-method.tex
\section{Method}
\label{sec:method}

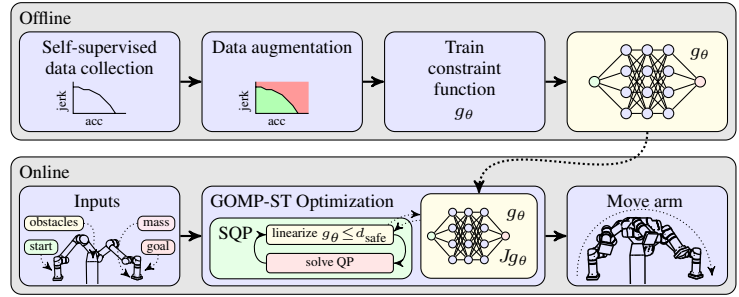
\begin{figure}[t]
    \centering
    \input{figures/pipeline}
    \caption{\textbf{\algname{} pipeline.}  In an offline process (\textbf{top}), \algname{} performs self-supervised learning of a constraint function.  It first repeatedly grasps objects of known mass to perform a boundary search on a motion profile parameterized by jerk and acceleration limits.  Then, during data augmentation, it labels slower motions as grasp successes, and faster motions as grasp failures.  Finally, it trains a neural network constraint function $g_\theta$.  In the online process (\textbf{bottom}), \algname{} computes a motion plan for a given problem.  The SQP solver repeatedly linearizes the learned suction constraint using a user-specified threshold $d_\mathrm{safe} \in [0,1]$ and the output and Jacobian (via autograd) of the trained neural network.}
    \label{fig:pipeline}
\end{figure}

\algname{} learns a suction cup acceleration constraint via a sequence of physical experiments, then integrates it with a time-optimizing motion planner.  See Fig.~\ref{fig:pipeline} for an overview.   This section starts with background on the prior work, then describes how to learn and integrate the constraint into the motion planner.  

\subsection{Background: GOMP-FIT}

The GOMP-FIT~\cite{ichnowski2022gompfit} algorithm formulates a fast-inertial-transport motion planning as an optimization problem and solves it with sequential quadratic program (SQP) trust-region-based solver.  It first discretizes the trajectory into a sequence of $H+1$ waypoints $(\vec{x}_0, \ldots \vec{x}_H)$ that are each separated by a fixed time interval $t_\mathrm{step}$.  Each waypoint $\vec{x}_t$ includes the configuration and its first and second derivatives $\vec{x}_t = (\vec{q}_t, \dot{\vec{q}}_t, \ddot{\vec{q}}_t)$.  The outer loop shrinks $H$ to find a minimum time trajectory.  The inner loop solves an SQP where the optimization objective minimizes the sum-of-squared accelerations, the linear constraints keep the motion within the kinematic and dynamics limits of the robot, and the non-linear constraints avoid obstacles and limit linear accelerations experienced at the end-effector.  To limit shock, it constrains the magnitude of end-effector acceleration to be below a threshold $a_\mathrm{safe}$.  To avoid spills, it constrains the angle between the container normal and the acceleration to be below a spill threshold $\theta_\mathrm{spill}$.  The optimization is
\begin{align*}
    \min_{\vec{x}_{[0..H]}} \quad & \frac 12 \sum_{t=0}^H \ddot{\vec{q}}_t \\
    \text{s.t.} \quad
    & \vec{x}_\mathrm{min} \le \vec{x}_t \le \vec{x}_\mathrm{max} \quad \forall t \in [0..H] \\
    & \vec{q}_{t+1} = \vec{q}_{t} + \dot{\vec{q}}_t t_\mathrm{step} + \frac{1}{2} \ddot{\vec{q}}_t t_\mathrm{step}^2 \quad \forall t \in [0..H) \\
    & \dot{\vec{q}}_{t+1} = \dot{\vec{q}}_t + \ddot{\vec{q}}_t t_\mathrm{step} \quad \forall t \in [0..H) \\
    & f_\mathcal{O}(\vec{q}) \ge 0 \quad \forall t \in [0..H]\\
    & \cos^{-1} (f_a(\vec{q}, \dot{\vec{q}}, \ddot{\vec{q}}) \cdot f_\mathrm{n}) \le \theta_\mathrm{spill} \quad \forall t \in [0..H] \\
    & \lVert f_a(\vec{q}, \dot{\vec{q}}, \ddot{\vec{q}}) \rVert \le a_\mathrm{safe} \quad \forall t \in [0..H],
\end{align*}
where $f_\mathcal{O} : \mathcal{C} \rightarrow \mathbb{R}$ is the signed distance from robot to set $\mathcal{O}$ (thus implementing the constraint $\vec{q}_t \in \mathcal{C}_\mathrm{free}$) and $f_a : (\mathcal{C})^3 \rightarrow \mathbb{R}^3$ is the linear acceleration at the end-effector computed using the Recursive Newton-Euler (RNE) algorithm. %
Additionally, GOMP-FIT optionally integrates constraints to optimize the grasp angle and location.

The SQP solver repeatedly linearizes the non-linear constraints (collision, end-effector acceleration, and grasp) to form a quadratic program (QP), and solves the QP, accepting solutions that improve the trajectory.  In this optimization, when solving for the $(k+1)$ iterate, a non-linear constraint of the form $g(\vec{x}) \le y$ is linearized around the current iterate $\vec{x}^{(k)}$ via a first-order approximation using its Jacobian:
\[
J \vec{x}^{(k+1)} \le y - g(\vec{x}^{(k)}) + J \vec{x}^{(k)}.
\]

GOMP-FIT optimizes the trajectory time by repeatedly solving the SQP with a shrinking horizon $H$, warm-starting each subsequent SQP solve with an interpolation of the solution from the prior horizon.  The optimization terminates the smallest $H$ the solver detects as feasible. However, when executed on robots with suction grippers, the resulting motions can lead to suction failures.

\subsection{Learned constraints in the SQP}
\input{figures/cup_deformation}

Maintaining a suction grasp throughout a high-speed motion requires balancing suction forces against inertial forces while the suction cups deform under strain. 
\algname{} defines a series of physical experiments to learn a constraint from real-world data.  We model the learned constraint as a function $g_\theta : (\mathcal{X})^h \rightarrow [0, 1]$ parameterized by $\theta$, where $(\mathcal{X})^h$ is history of $h$ dynamic states, and a value of 0 indicates the object is held, while a value of 1 indicates a failure.  (Notationally, we use $h$ to indicate a history of states, and $H$ to indicate the total trajectory length.)

We integrate this function into the optimization as another non-linear constraint in the form:
\[
g_\theta(\cdot) \le d_\mathrm{safe},
\]
where the argument is a portion of the state from of the trajectory being optimized, and $d_\mathrm{safe} \in [0,1)$ is a tunable failure threshold.  To linearize this constraint for the SQP, we use the automatic gradients (Autograd) provided by a neural-network package~\cite{bradbury2021jax} to compute the Jacobian.

Using slow-motion video capture, we observe that suction cups deform before suction failure (Fig.~\ref{fig:cup_deformation}), and thus failures are not an instantaneous response to a change in end-effector state. %
With this observation, we propose that $g_\theta$ should be a function of a sequence (or history) of states of length $h$.
The value of $h$ depends on the geometry and material of the suction cup.  In a series of experiments we set $h$ to be long enough to capture the time between deformation start and suction failure that we observe from slow-motion video playback ($>$ 0.1 seconds).
Further, to translate the optimization variables from the state of the robot to the state of the end-effector, we utilize RNE function $f_a$ for each state in the history.
Thus we formulate the full constraint as: %
\[
g_\theta(f_a(\vec{x}_{t-h+1}), \ldots, f_a(\vec{x}_{t-1}), f_a(\vec{x}_{t})) \le d_\mathrm{safe} \quad \forall t \in [1 .. H).
\]
When $\vec{x}$ has a negative subscript (i.e., it refers to a state before the start of planning), we replace it with $\vec{x}_0$. 

\subsection{Self-supervised data collection and training}
\label{sec:training}

To learn $g_\theta$, \algname{} implements a self-supervised pipeline that defines a series of experiments to lift objects of known mass and collect data. %
We attach a pressure sensor to the tube connected to the suction cups, similar to Huh et al.~\cite{huh2021multi}.  To minimize delay between pressure changes and pressure readings, we place the sensor close to the suction cups.  During each lift, the pipeline records the joint state and pressure sensor readings over time.

To isolate gravity during data collection, the system always lifts vertically while varying the angle of the suction normal relative to gravity (see  Fig.~\ref{fig:cup_deformation}).  As convention, $0^\circ$ is a top-down grasp and $90^\circ$ is a grasp in which the suction normal is perpendicular to gravity.

The pipeline computes each lift motion using Ruckig~\cite{berscheid2021jerk} between two points in end-effector (Cartesian) coordinates, and uses an inverse kinematics solver to translate into joint configurations.  Ruckig computes straight-line time-optimal motions subject to velocity, acceleration, and jerk limits.  During data collection, \algname{} varies the motion profile by changing the maximum acceleration $a_\mathrm{max}$ and maximum jerk $j_\mathrm{max}$ parameters of Ruckig.

The pipeline defines a discretized grid with $a_\mathrm{max}$ and $j_\mathrm{max}$ axes to fill with values 0 or 1 labels.  To reduce data collection time, \algname{} performs a boundary search on the motion profiles.  It starts with a fixed lower value of $a_\mathrm{max}$ and increases $j_\mathrm{max}$ until it observes a change in the pressure measurements indicating a suction grasp failure.  Afterwards, it iteratively decreases $j_\mathrm{max}$ or increases $a_\mathrm{max}$ so that it is always exploring the above and below the continuous boundary at which suction fails.

After a suction failure, the automated data collection pipeline takes a top-down image of the scene to find and re-grasp the target object. After confirming the grasp using the pressure measurements, \algname{} moves the object to a consistent starting pose before performing the next lift. %

The pipeline then trains a multi-layer perceptron with exponential linear units (ELU) activations using the joint and pressure data it collected.  We choose ELU for its continuous gradients.  Training details are in the experiments section.  The pipeline scans the recorded pressure data to find the time at which suction pressure is lost.  It then tracks a tunable number of steps $h$ back to create a labeled data point containing $h$ accelerations that led to the suction failure (labelled as 1).  All sequences of $h$ accelerations prior to the suction failure, or in records without suction failures, the pipeline labels as 0.  \algname{} further perform data augmentation by scaling accelerations leading to failures by $1 + \epsilon$, and non-failures by $1 - \epsilon$, for small positive values of $\epsilon$.

\subsection{Analytic model of suction-cup failure for GOMP-FIT baseline}
\label{sec:analytic}
As a baseline to the learned model, we compare \algname{} to GOMP-FIT where its constraint on the magnitude and direction of the inertial acceleration vector is provided by an analytic model. In previous work, Kolluru et al.~\cite{kolluru1998} and Stuart et al.~\cite{Stuart2015SuctionHI} proposed analytic models for rigid suction cups, but were limited to either single suction cups or symmetrical systems. Our analysis is more closely related to the more general method used by Valencia et al.~\cite{valencia2017} and Pham and Pham~\cite{pham2019critically} which generalises over multiple suction cups and asymmetric loads, though still with the assumption of rigid bodies. We also choose the rigid-body assumption as opposed to for example the spring model suggested by Mahler et al.~\cite{mahler2018dex}, since a spring model of suction cups where quasi-static equilibrium is not assumed would require knowledge about the state of the springs. Alternatively, if quasi-static equilibrium is assumed, the state of the springs may be estimated, but the purpose of the additional complexity is defeated from a motion planning perspective as the resulting constraint would be equivalent to that of a rigid-body model. %
In this work, the analytic model makes the following assumptions:
\begin{enumerate}
    \item There are quasi-static conditions in the inertial frame, enabling equilibrium analysis.
    \item The suction cups are rigid and modelled as point contacts.
    \item The transported object is a rigid rectangular cuboid with uniform mass distribution such that its center of mass corresponds to its centroid.
    \item There are no air leaks between the suction cups and flat grasping surface, which results in a static and equal suction force across all suction cups.
\end{enumerate}

Consider the free-body diagram in the inertial frame of the object shown in Fig.~\ref{fig:fbd}, where $\vec{f}_{s,i}$ are the suction forces, $\vec{f}_{\vec{n},i}$ are the reaction/contact forces between the suction cups and grasped object, $\vec{f}_{f,i}$ are the friction forces, $\vec{f}_g$ is the gravitational force, and $\vec{a}$ is the acceleration resulting from a balance of forces. $\vec{\hat{n}}_\mathrm{obj}$ is the unit length (denoted by the hat) normal defining the grasping plane, and $\theta = \cos^{-1} \left[ (\vec{a}\cdot \vec{\hat{n}}_\mathrm{obj}) / (||\vec{a}||_2 ||\vec{\hat{n}}_\mathrm{obj}||_2) \right]$ is the angle between the inertial acceleration vector and suction force normal. In the idealised condition, $\vec{f}_{s,i} = (p_\mathrm{atm} - p_v)A$, where $p_\mathrm{atm} - p_v$ is the difference between atmospheric pressure $p_\mathrm{atm}$ and applied vacuum pressure $p_v$, and $A$ is the effective area of grip for a single suction cup. Furthermore, under the assumption of dry Coulomb friction, $\vec{f}_{f,i} \leq \mu \vec{f}_{\vec{n},i}$, where $\mu$ is the static coefficient of friction between the grasped object and suction cups. Finally, $\vec{f}_g = m \vec{g}$, where $m$ is the object mass, and $\vec{g}$ is gravitational acceleration.

\input{figures/FBD}

We define 2 models, \emph{unimodal} and \emph{multimodal}.  The unimodal model assumes %
$\theta = 0^\circ$, and thus %
the maximum inertial acceleration before grasp failure is trivially given by $\vec{a}_{\mathrm{fail}} = \frac{1}{m} \sum_i \vec{f}_{s,i}$. %
The multimodal model includes multiple failure modes for cases where $\theta \neq 0^\circ$, including
suction grasp failure by sliding, force imbalance, or moment imbalance.  This model does not include deformation.  The analysis uses the equations %
\begin{align}
    \label{eq:force_balance}
    \sum_i \vec{f}_{s,i} + \sum_i \vec{f}_{\vec{n},i} + \sum_i \vec{f}_{f,i} + \vec{f}_g = m \vec{a} \\
    \label{eq:moment_balance}
    \sum_i \vec{r}_i \times \vec{f}_{s,i} + \sum_i \vec{r}_i \times \vec{f}_{\vec{n},i} + \sum_i \vec{r}_i \times \vec{f}_{f,i} = \vec{0} \\
    \label{eq:reaction_direction}
    \vec{f}_{\vec{n},i} = -\alpha_i \cdot \vec{\hat{n}}_\mathrm{obj} \\
    \label{eq:friction_projection}
    \hat{\vec{f}}_{f,i} = -\widehat{\mathbf{proj}_{\vec{\hat{n}}_\mathrm{obj}} (\vec{f}_g)} \\
    \label{eq:friction_direction}
    \vec{f}_{f,i} = \beta_i \cdot \hat{\vec{f}}_{f,i} \\
    \label{eq:frcition_magnitude}
    (\vec{f}_{f,i} \cdot \hat{\vec{f}}_{f,i}) = \gamma \cdot \vec{f}_{\vec{n},i} \cdot (-\vec{\hat{n}}_\mathrm{obj}).
\end{align}
The force (\ref{eq:force_balance}) and moment (\ref{eq:moment_balance}) balance in quasi-static equilibrium,
where $\vec{r}_i$ is the position vector from the center of mass of the object to the point of application of forces for the $i$th suction cup. We then require the reaction forces $\vec{f}_{\vec{n},i}$ to be normal to the grasping plane (\ref{eq:reaction_direction}), set the direction of the friction forces to be tangent in the grasping plane and opposing the direction of the motion in its absence (\ref{eq:friction_projection}), (\ref{eq:friction_direction}), and finally, require that the friction force magnitudes share the same proportionality constant $\gamma$ to the normal forces (\ref{eq:frcition_magnitude}).
 
We solve the system of equations for a range of hypothetical angles $\theta$ and accelerations $\vec{a}$, and classify each scenario as a failure or a success based on the physical restrictions that $\beta_i \leq \mu\alpha_i$ and $\alpha_i \geq 0$, where $\alpha_i$ and $\beta_i$ are the magnitudes of the normal and frictional forces, respectively. The result of this simulation is shown in Fig.~\ref{fig:a_v_theta}, which shows that the multimodal model converges with the unimodal model when $\theta \rightarrow 0^+$, and that in contrast to intuition, the curve is not strictly increasing nor strictly decreasing across the domain. The inputs used to generate the curve use dimensions from our experimental setup, and so the exact axes values do not generalise to other systems. To integrate the model with GOMP-FIT, we approximate the curve using a 4th-order polynomial fit, which we then use to formulate an analytic baseline constraint.

\input{figures/a_v_theta}

%% file: figures/pipeline.tex
\begin{tikzpicture}[
    >=stealth',
    font=\scriptsize,
    baseblock/.style={
        draw, rectangle, rounded corners, inner sep=3pt},
    baselabel/.style={
        inner sep=3.2pt, execute at end node={\vphantom{Ig}}},
    topblock/.style={
        baseblock, fill=black!10, minimum width=(3.8in - 0.4pt), minimum height=52pt},
    toplabel/.style={
        baselabel, below=0pt of #1.north west, align=left, anchor=north west},
    flowblock/.style={
        baseblock, fill=blue!10},
    flowlabel/.style={
        baselabel, below=0pt of #1.north, anchor=north, align=center},
    leftlabel/.style={
        baselabel, below=0pt of #1.north west, anchor=north west, align=left},
    neuron/.style={
        circle, draw=black, fill=blue!10, inner sep=0, minimum size=4pt},
    neuronsm/.style={neuron, minimum size=3pt},
    minibox/.style={baselabel, minimum width=48pt, rounded corners=1.4pt, inner sep=1pt, font=\tiny, draw=black},
    inputbox/.style={inner xsep=2pt, inner ysep=0.5pt, draw, very thin, rounded corners=2pt}
]
\node [topblock] (offline) {};
\node [toplabel={offline}] (offline label) {Offline};
\node [topblock, below=6pt of offline] (online) {};
\node [toplabel={online}] (online label) {Online};

\draw let \p1=(offline label.south west),
          \p2=(offline.south east),
          \n1={\y1 - \y2 - 4*0.4pt}, %
          \n2={8pt}, %
          \n3={(\x2 - \x1 - 8pt - \n2*3)/4} %
    in
    node [flowblock, below=-2pt of offline label.south west, anchor=north west, xshift=3.2pt, minimum height=\n1, minimum width=\n3] (collect) {}
    node [flowlabel=collect] (collect label) {Self-supervised \\ data collection}
    node [flowblock,right=\n2 of collect, minimum height=\n1, minimum width=\n3] (augment) {}
    node [flowlabel=augment] {Data augmentation}
    node [flowblock,right=\n2 of augment, minimum height=\n1, minimum width=\n3] (train) {}
    node [flowlabel=train] {Train \\ constraint \\ function \\ $g_\theta$}
    node [flowblock,right=\n2 of train, minimum height=\n1, minimum width=\n3, fill=yellow!10] (nn off) {}
    node [flowblock, below=-2pt of online label.south west, anchor=north west, xshift=3.2pt, minimum height=\n1, minimum width=\n3] (inputs) {}
    node [flowlabel=inputs] (inputs label) { Inputs } %

    node [flowblock, right=\n2 of inputs, minimum height=\n1, minimum width={\n3*2 + \n2 + 0.4pt}] (opt) {}
    node [leftlabel=opt] (opt label) { GOMP-ST Optimization }
    node [flowblock, right=\n2 of opt, minimum height=\n1, minimum width=\n3] (move) {}
    node [flowlabel=move] { Move arm };

\coordinate [yshift=2pt, xshift=-10pt] (collect graph) at (collect label.south);
\path [draw] (collect graph) -- ++(0,-12pt) -- +(20pt,0)
   +(0,10pt) -- +(2pt,10pt) -- +(4pt,9pt) -- +(6pt, 9pt) -- +(8pt, 8pt) -- +(10pt, 7pt) -- +(12pt, 5pt) -- +(14pt, 3pt) -- +(16pt, 0pt);
\node [left=2pt of collect graph, rotate=90, inner sep=0, xshift=-3pt, yshift=2pt] { \tiny jerk };
\node [below=14pt of collect graph, inner sep=0, xshift=8pt] { \tiny acc };

\coordinate [xshift=-10pt] (augment graph) at (collect graph -| augment.south);
\coordinate [yshift=-12pt] (augment origin) at (augment graph);
\path [fill=green!30] (augment origin) %
   -- +(0,10pt) -- +(2pt,10pt) -- +(4pt,9pt) -- +(6pt, 9pt) -- +(8pt, 8pt) -- +(10pt, 7pt) -- +(12pt, 5pt) -- +(14pt, 3pt) -- +(16pt, 0pt);
   
\path [fill=red!40] (augment origin) %
   +(20pt,0) -- +(20pt,12pt) -- +(0pt,12pt)
   -- +(0,10pt) -- +(2pt,10pt) -- +(4pt,9pt) -- +(6pt, 9pt) -- +(8pt, 8pt) -- +(10pt, 7pt) -- +(12pt, 5pt) -- +(14pt, 3pt) -- +(16pt, 0pt);
   
\path [draw] (augment graph) -- ++(0,-12pt) -- +(20pt,0)
+(0,10pt) -- +(2pt,10pt) -- +(4pt,9pt) -- +(6pt, 9pt) -- +(8pt, 8pt) -- +(10pt, 7pt) -- +(12pt, 5pt) -- +(14pt, 3pt) -- +(16pt, 0pt);

\node [left=2pt of augment graph, rotate=90, inner sep=0, xshift=-3pt, yshift=2pt] { \tiny jerk };
\node [below=14pt of augment graph, inner sep=0, xshift=8pt] { \tiny acc };

\path [thick, ->]
    (collect) edge (augment)
    (augment) edge (train)
    (train) edge (nn off)
    (inputs) edge (opt)
    (opt) edge (move);
    
\node [above=1pt of inputs.south, anchor=south, inner sep=0] { \includegraphics[width=40pt]{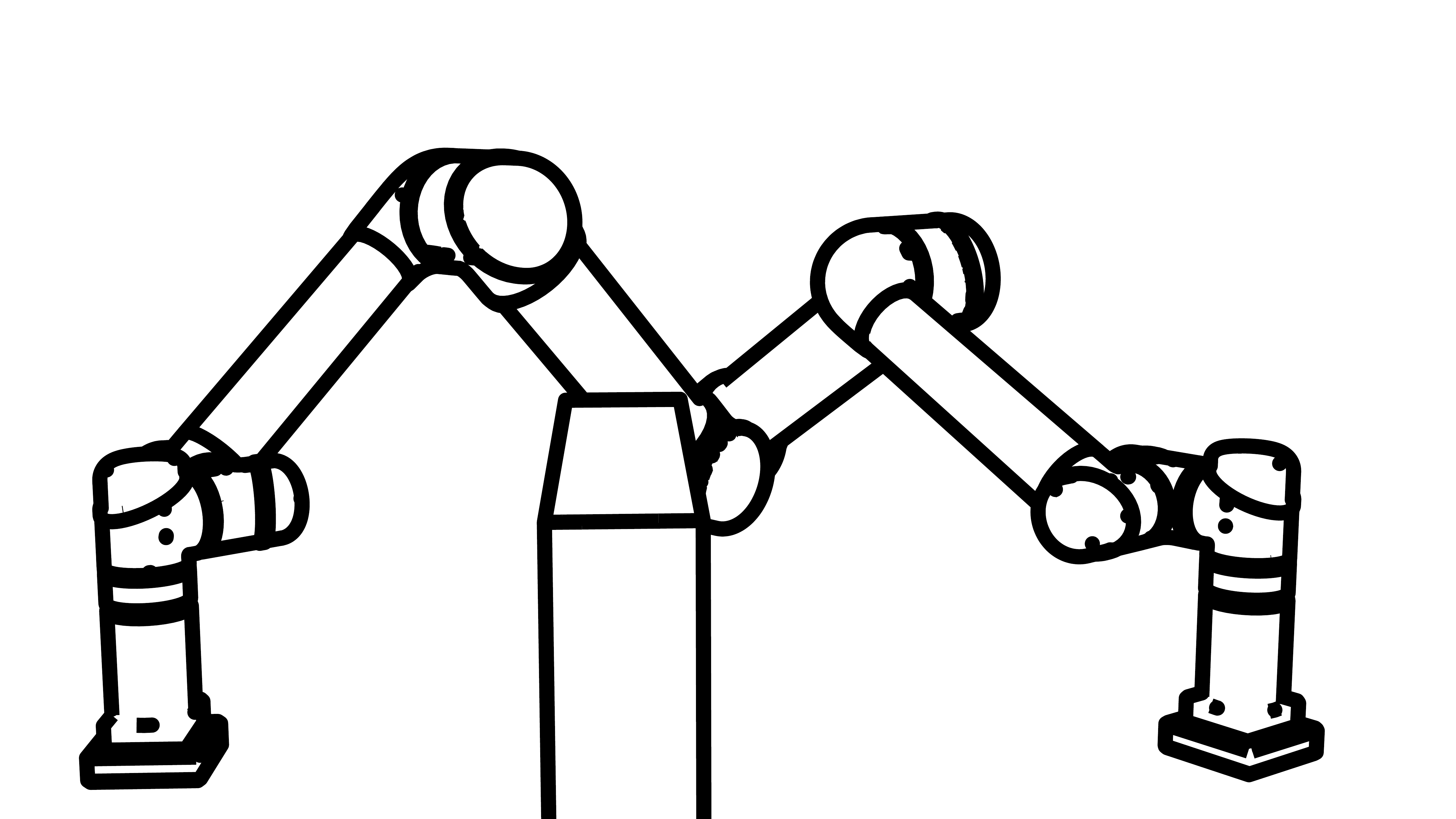}};

\node [inputbox, above=12pt of inputs.south west, anchor=south west,  xshift=2pt, fill=green!10] (start input label) {\tiny start\vphantom{gl}};
\node [inputbox, above=12pt of inputs.south east, anchor=south east,  xshift=-2pt, fill=red!10] (goal input label) {\tiny goal };
\node [inputbox, above=21pt of inputs.south west, anchor=south west, xshift=2pt, fill=yellow!10] (obstacle input label) { \tiny obstacles\vphantom{gl} };
\node [inputbox, above=21pt of inputs.south east, anchor=south east, xshift=-2pt, fill=magenta!10] (mass input label) { \tiny mass\vphantom{gl} };

\coordinate [xshift=-18pt, yshift=6pt] (start input point) at (inputs.south);
\coordinate [xshift=17pt, yshift=6pt] (goal input point) at (inputs.south);
\coordinate [yshift=10.5pt, xshift=-2.5pt] (obstacle input point) at (inputs.south);
\coordinate [xshift=11.5pt, yshift=3.4pt] (mass input point) at (inputs.south);

\draw [densely dotted, ->] (start input label.south) .. controls +(270:0.1) and +(140:0.2) .. (start input point);

\draw [densely dotted, ->] (goal input label.south) .. controls +(270:0.1) and +(30:0.2) .. (goal input point);

\draw [densely dotted, ->] (obstacle input label.east) .. controls +(0:0.2) and +(90:0.2) .. (obstacle input point);

\draw [densely dotted, ->] (mass input label.west) .. controls +(180:0.2) and +(140:0.4) .. (mass input point);

\node [above=1pt of move.south, anchor=south, inner sep=0] {
\includegraphics[width=52pt]{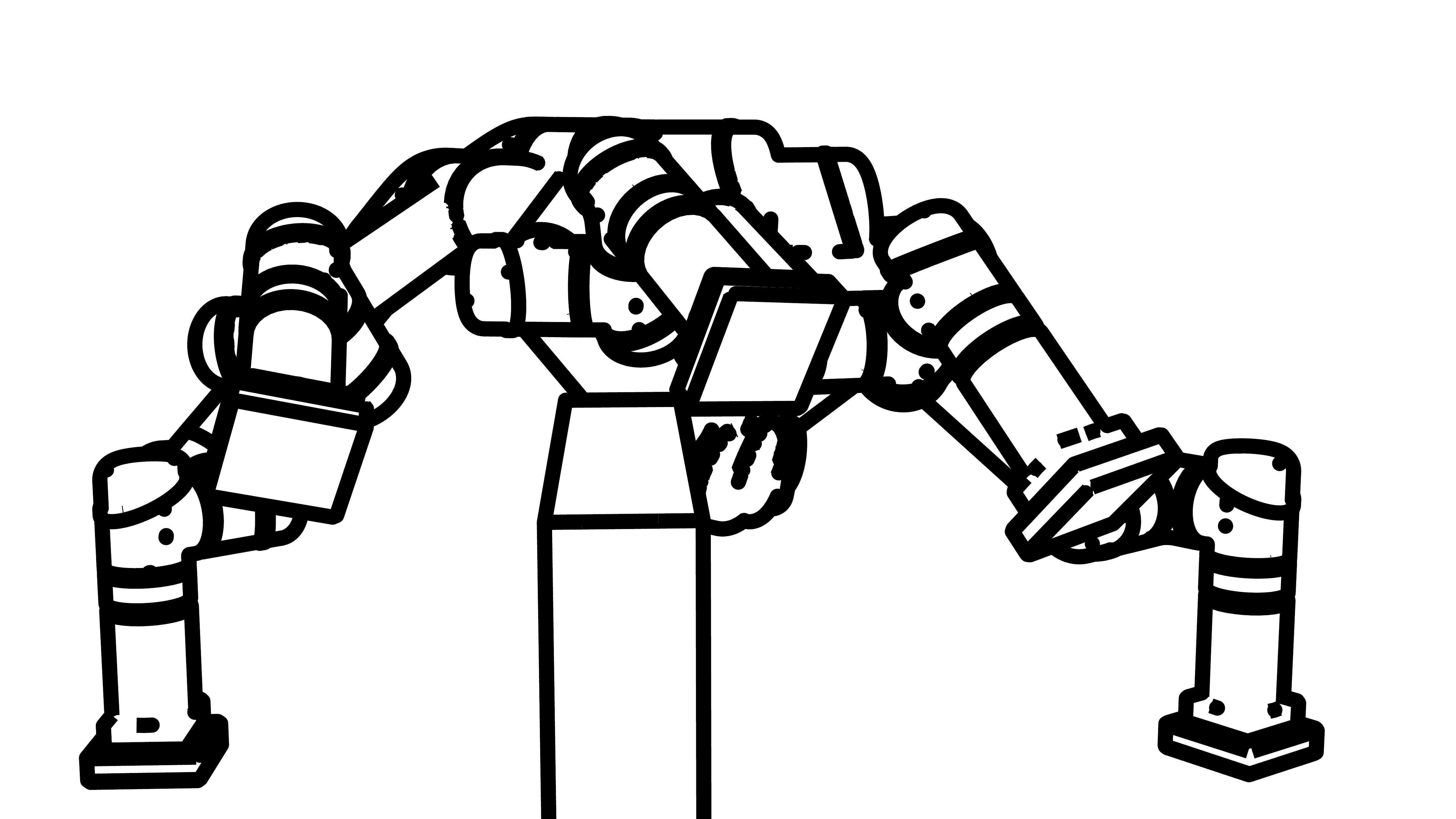}};

\draw [densely dotted, ->] ($(move.south west) + (4pt,4pt)$) .. controls +(90:1.2) and +(95:1.15) .. ($(move.south east) + (-6pt,4pt)$);

\node [neuron, xshift=-20pt, fill=green!10] (nn off in) at (nn off.center) {};
\node [neuron, xshift=20pt, fill=red!10] (nn off out) at (nn off.center) {};
\foreach \i in {0,1,2,3}{
    \node [neuron, xshift=-8pt, yshift={(\i-1.5)*8pt}] (nn off 1 \i) at (nn off.center) {};
    \node [neuron, xshift=0pt, yshift={(\i-1.5)*8pt}] (nn off 2 \i) at (nn off.center) {};
    \node [neuron, xshift=8pt, yshift={(\i-1.5)*8pt}] (nn off 3 \i) at (nn off.center) {};
    \draw [thin] (nn off in) -- (nn off 1 \i)
          (nn off 3 \i) -- (nn off out);
}
\foreach \i in {0,1,2,3}{
    \foreach \j in {0,1,2,3}{
        \draw [thin] (nn off 1 \i) -- (nn off 2 \j);
        \draw [thin] (nn off 2 \i) -- (nn off 3 \j);
    }}
    
\node [above=3pt of nn off out] {$g_\theta$};

\coordinate [xshift=1.4pt] (nn on) at (train.center |- opt.center);
\node [flowblock, draw=black, minimum height=32pt, minimum width=44pt, xshift=4pt, fill=yellow!10] (nn on box) at (nn on) { };
\node [neuronsm, xshift=-14pt, fill=green!10] (nn on in) at (nn on.center) {};
\node [neuronsm, xshift=14pt, fill=red!10] (nn on out) at (nn on.center) {};
\foreach \i in {0,1,2,3}{
    \node [neuronsm, xshift=-6pt, yshift={(\i-1.5)*6pt}] (nn on 1 \i) at (nn on.center) {};
    \node [neuronsm, xshift=0pt, yshift={(\i-1.5)*6pt}] (nn on 2 \i) at (nn on.center) {};
    \node [neuronsm, xshift=6pt, yshift={(\i-1.5)*6pt}] (nn on 3 \i) at (nn on.center) {};
    \draw [thin] (nn on in) -- (nn on 1 \i)
          (nn on 3 \i) -- (nn on out);
}
\foreach \i in {0,1,2,3}{
    \foreach \j in {0,1,2,3}{
        \draw [thin] (nn on 1 \i) -- (nn on 2 \j);
        \draw [thin] (nn on 2 \i) -- (nn on 3 \j);
    }}
    
\node [above left=2pt of nn on out, xshift=14pt] {$g_\theta$};
\node [below left=1pt of nn on out, xshift=14pt] {$J\!g_\theta$};

\draw [densely dotted, thick, ->] (nn off.south) .. controls +(270:1) and +(90:1) .. (nn on box.north);

\coordinate [xshift=6.4pt, yshift=-2.4pt] (opt nw) at (opt label.south west);
\coordinate [xshift=-6.4pt, yshift=3.2pt] (opt sw) at (nn on box.south west);
\node [flowblock, fit=(opt nw)(opt sw), inner sep=0, align=left, inner sep=3.2pt, fill=green!10] (sqp) { };
\node [leftlabel={sqp}] (sqp label) { SQP };

\node [minibox, right=0pt of sqp label, fill=yellow!10] (linearize) { linearize $g_\theta{\le}d_\text{safe}$}; 
\node [minibox, below=3pt of linearize, fill=red!10] (solve qp) { solve QP\vphantom{$g_\theta{\le}d_\text{safe}$} };

\draw [->] (linearize.east) .. controls +(0:0.2) and +(0:0.2) .. (solve qp.east);
\draw [->] (solve qp.west) .. controls +(180:0.2) and +(180:0.2) .. (linearize.west);

\path [draw, densely dotted] (linearize.10) edge [bend left=20, ->] (nn on box.160);

\path [draw, densely dotted] (linearize.3) edge [bend left=20, <-] (nn on box.165);

\end{tikzpicture}

%% file: figures/cup_deformation.tex
\begin{figure}[t]
    \centering

    \begin{tabular}{@{}cc@{}}
        \begin{tabular}{@{}c@{}}
            \includegraphics[height=39pt]{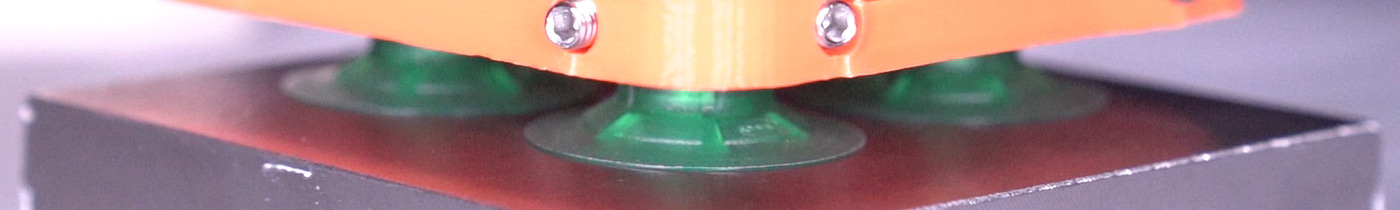} \\[0pt]
            \includegraphics[height=39pt]{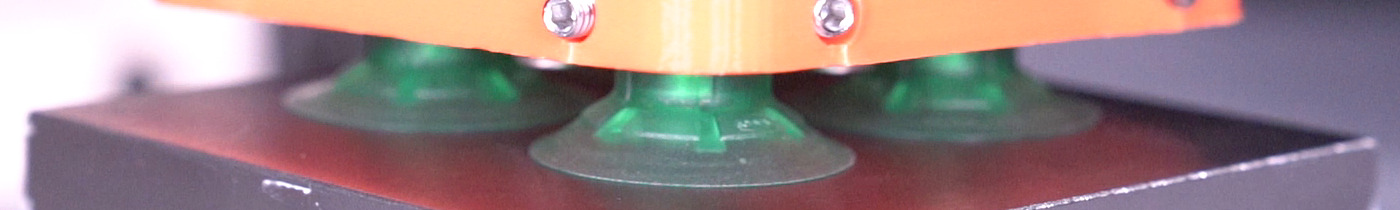} \\[0pt]
            \includegraphics[height=39pt]{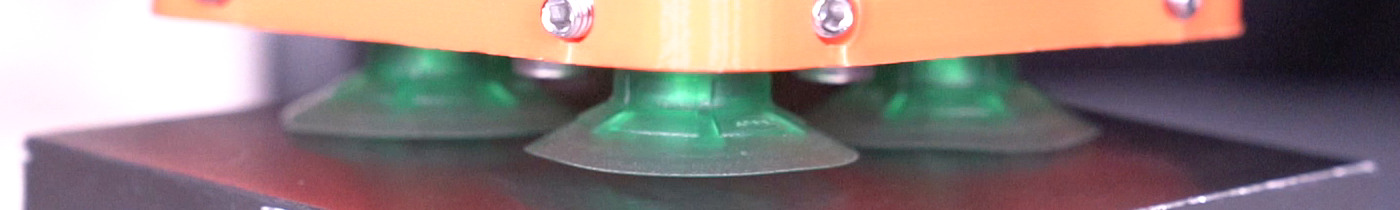}
        \end{tabular}
        &

        \begin{tabular}{@{}c@{}}
         \includegraphics[height=124pt,clip,trim={0 250 0 0}]{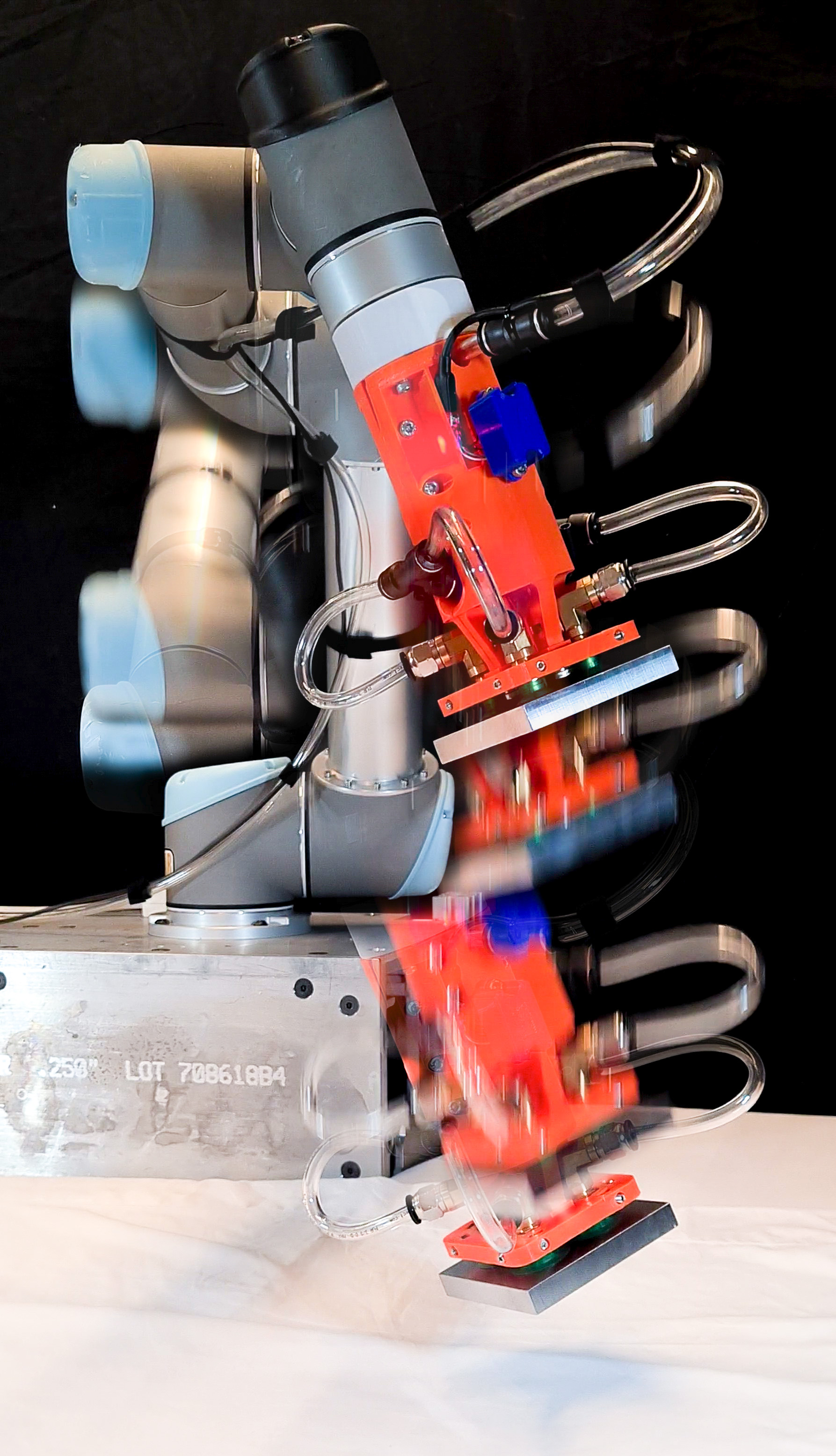} \\
          \end{tabular} \\
(a) Suction cup deformation before and during suction break. & (b) Vertical lift motion%
    \end{tabular}

    \caption{\textbf{Suction cup deformation observed with data collection.}
    (a) These frames from a slow-motion video show the deformation of suction cups as the gripper is rapidly pulled upward. 
    At the beginning of the motion ((a) \textbf{top}), the suction cups seals and compresses against the grasped surface.  As the gripper moves up and starts to break from suction ((a) \textbf{middle}), we observe that the suction cups deform but still maintain a seal. Continuing to pull away results in a suction grasp failure ((a) \textbf{bottom}).
    (b) During data collection, the robot lifts the mass with a vertical motion while grasping at an angle.}
    \label{fig:cup_deformation}
\end{figure}

%% file: figures/FBD.tex
\begin{figure}[t]
    \centering
    \includegraphics[width=0.4\columnwidth]{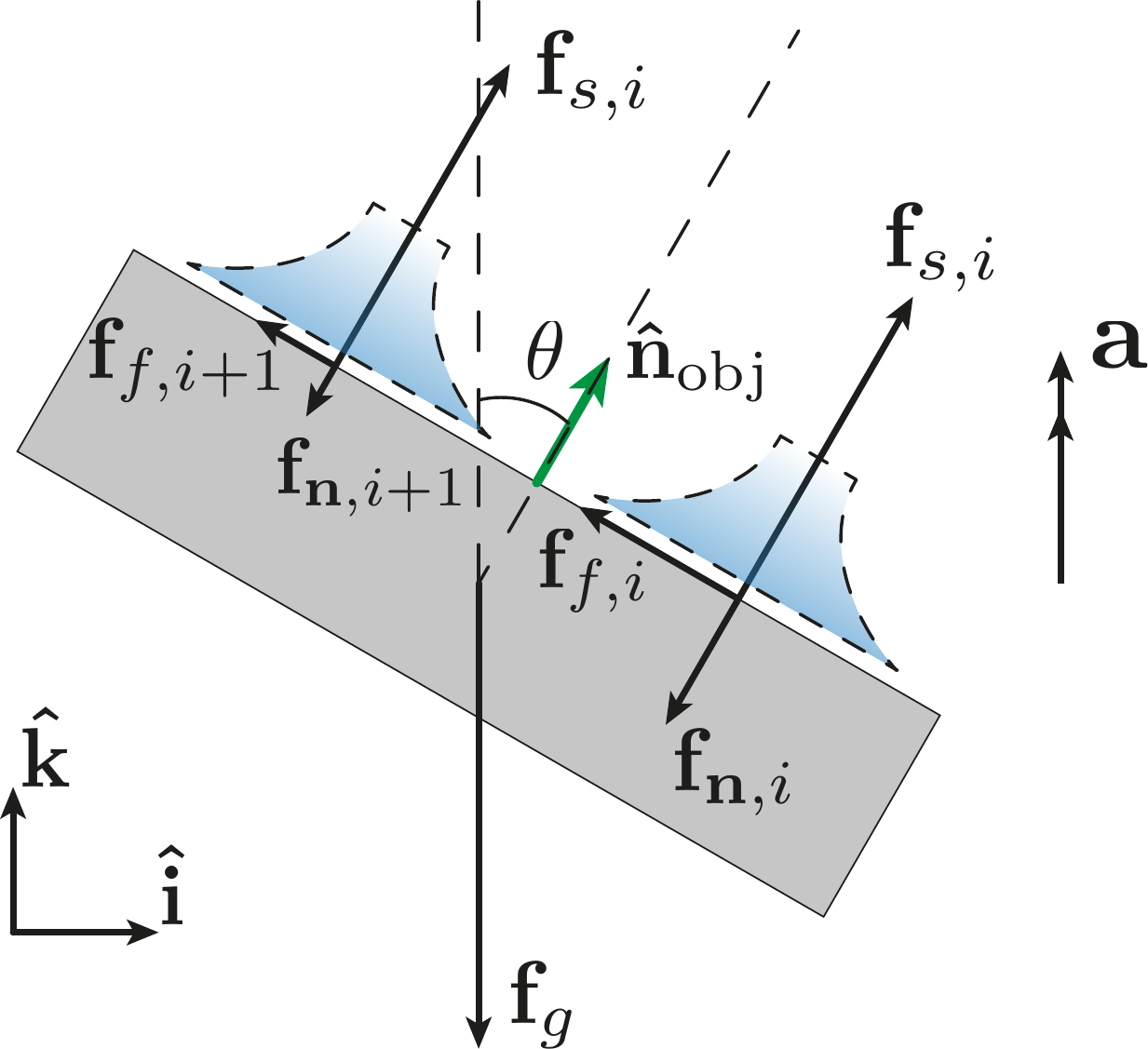} \\

    \caption{\textbf{Analytic model of suction cup failure}
    The gray block is the object held by a set of suction cups (light blue), where the suction surface normal is defined by vector $\hat{\vec{n}}$ (not shown), which is always aligned with vector $\hat{\vec{n}}_\mathrm{obj}$.  The suction forces $\Vec{f}_{s,i}$ apply the same suction force at each suction contact force, while the normal forces $\Vec{f}_{\Vec{n},i}$ react unevenly based on the tilt angle $\theta$ relative to inertial acceleration and gravity $\Vec{f}_{g}$.
    The green arrow indicates the unit normal direction vector, and all other arrows are force vectors.
    Under the unimodal model $\theta = 0^\circ$. Under the multimodal model $\theta$ can vary.
    }
    \label{fig:fbd}
\end{figure}

%% file: figures/a_v_theta.tex
\begin{figure}[t]
    \centering
    \includegraphics[width=\columnwidth]{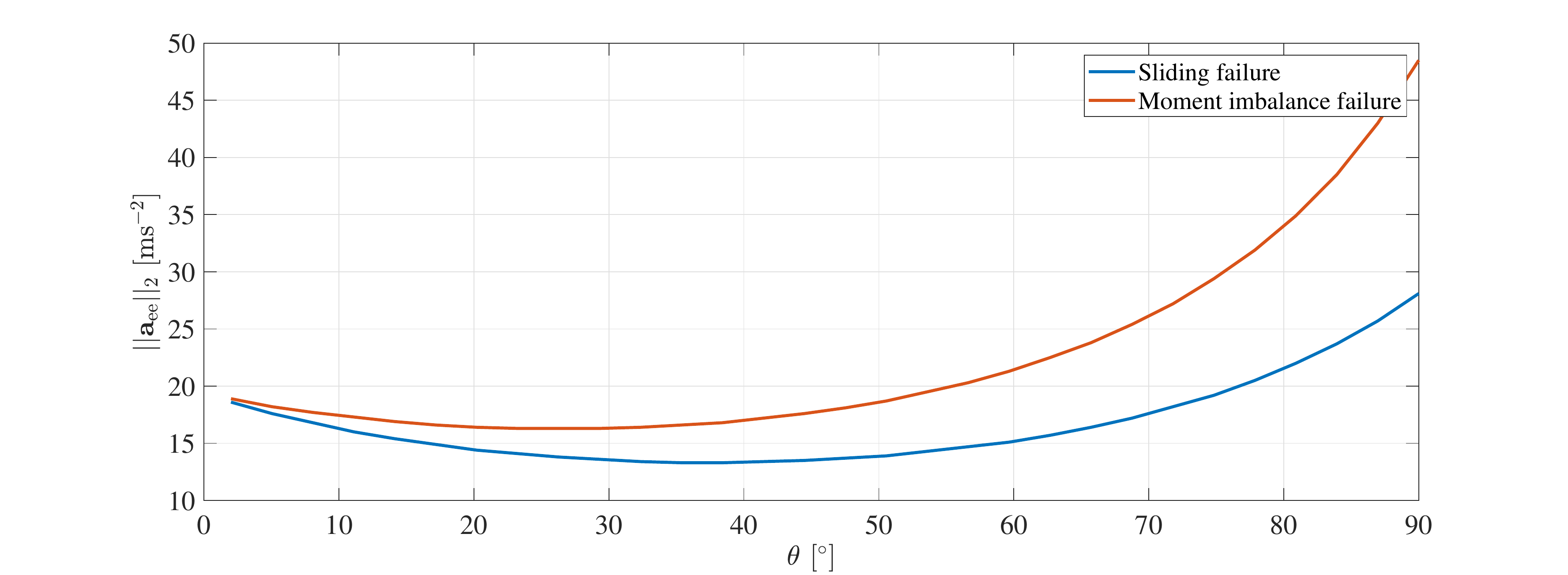} \\

    \caption{\textbf{Theoretical acceleration magnitude profile for the multimodal model.}
    The two lines are the acceleration magnitude drop boundaries in the sliding and moment imbalance failure modes for various angles $\theta$ between the inertial acceleration $\Vec{a}$ and the suction normal $\Vec{\hat{n}}_\mathrm{obj}$. According to the model, suction fails for accelerations in the region above the blue curve. We generate this plot by solving the system of equations (\ref{eq:force_balance}) -- (\ref{eq:frcition_magnitude}) with a combination of angles and accelerations, where grasp failures may occur by sliding, force imbalance (only when $\theta = 0$), or moment imbalance. For the parameters such as object dimensions and friction coefficient of our system, sliding will always occur before moment imbalance.}
    \label{fig:a_v_theta}
\end{figure}

%% file: section-5-experiments.tex
\section{Experiments}
\label{sec:experiments}
\input{figures/gripper_annotated}
\input{table/experiments_mass}

We perform experiments on a physical UR5 with a custom 4-cup vacuum gripper and a set of steel blocks.  The gripper has four round 30\,mm diameter elastodur flat suction cups driven by a single VacMotion MSV 27 vacuum generator.  Multi-cup suction grippers are common in many commercial automated logistics systems due to the increased surface area producing more suction force, and the multiple contact point stabilizing the hold resulting in reduced payload swing.  The pressure sensor (Adafruit MPRLS Ported Pressure Sensor Breakout) is fitted in the gripper assembly and attached via USB to the computer that drives the UR5.  See Fig.~\ref{fig:figure_one} for a visual of the experimental setup and Fig.~\ref{fig:gripper} for a close-up of the suction gripper.

\begin{figure}[t]
    \centering
    \begin{tikzpicture}[>=stealth',font=\scriptsize]
    \node [inner sep=0] (motion) {%
    \includegraphics[height=72pt,clip,trim=80 30 125 115]{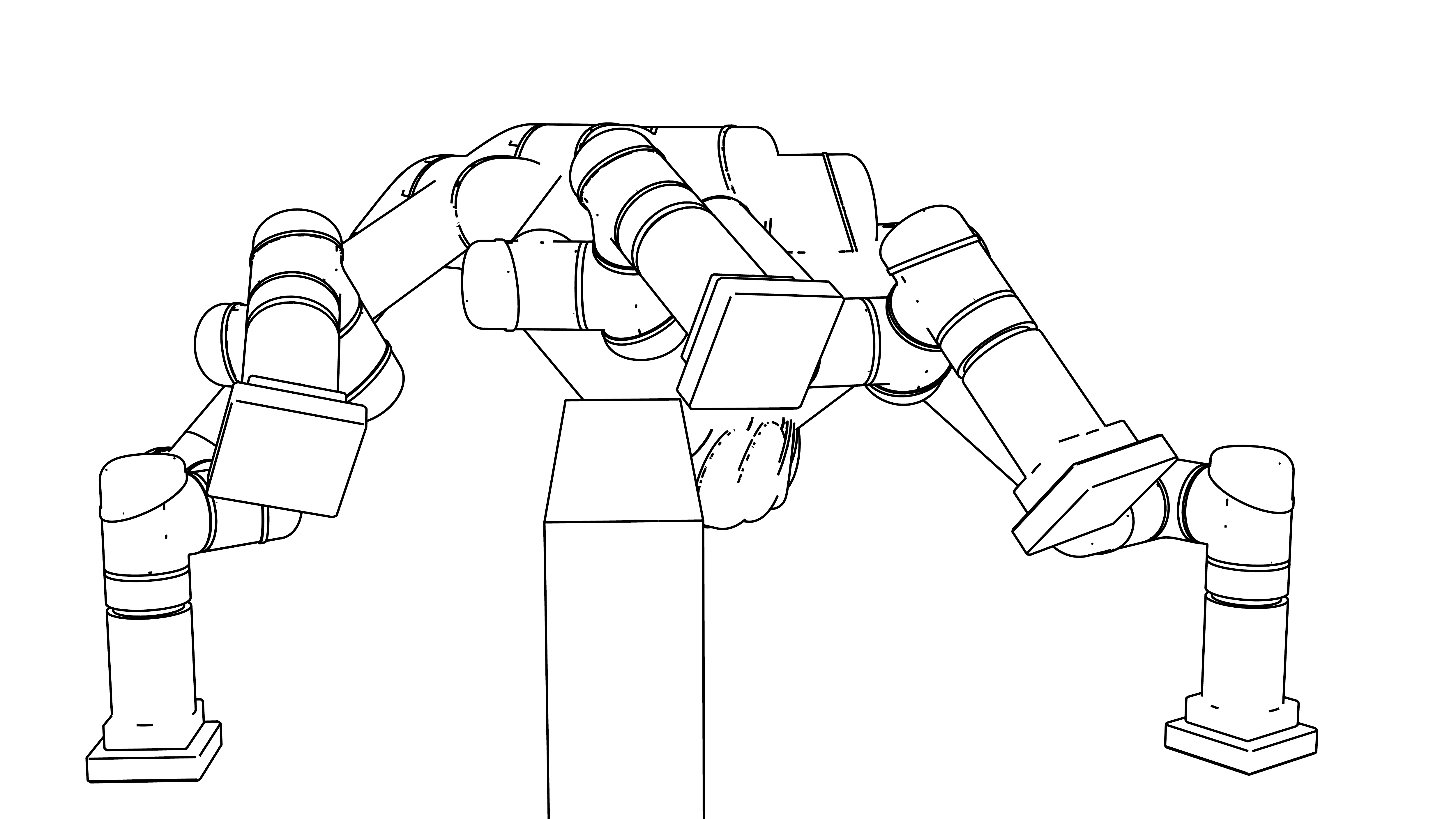}};
    \coordinate [xshift=8pt, yshift=-3pt] (motion start) at (motion.south west);
    \coordinate [xshift=-8pt, yshift=-3pt] (motion end) at (motion.south east);
    \draw [] ($(motion start)+(0,3pt)$) -- +(0, -6pt);
    \draw [] ($(motion end)+(0,3pt)$) -- +(0, -6pt);
    \draw [<->] (motion start) -- (motion end) node [midway, fill=white, inner sep=2pt] { horizontal distance };
    
    \node [above=6pt of motion.south, anchor=south west, inner sep=1pt, xshift=6pt] (obstacle label) { obstacle\vphantom{g} };
    \draw [->] (obstacle label.west) -- +(-12pt, 0pt);
    
    \node [right=of motion.north east, anchor=north west, inner sep=0] (mass a) {
        \includegraphics[height=20pt]{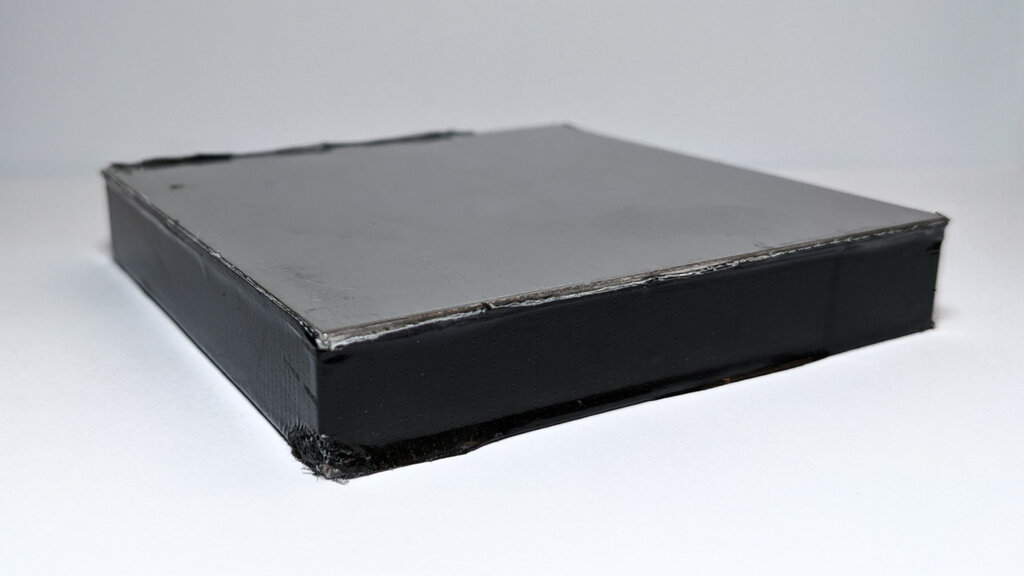}};
    \node [below=4pt of mass a, inner sep=0] (mass b) {
        \includegraphics[height=20pt]{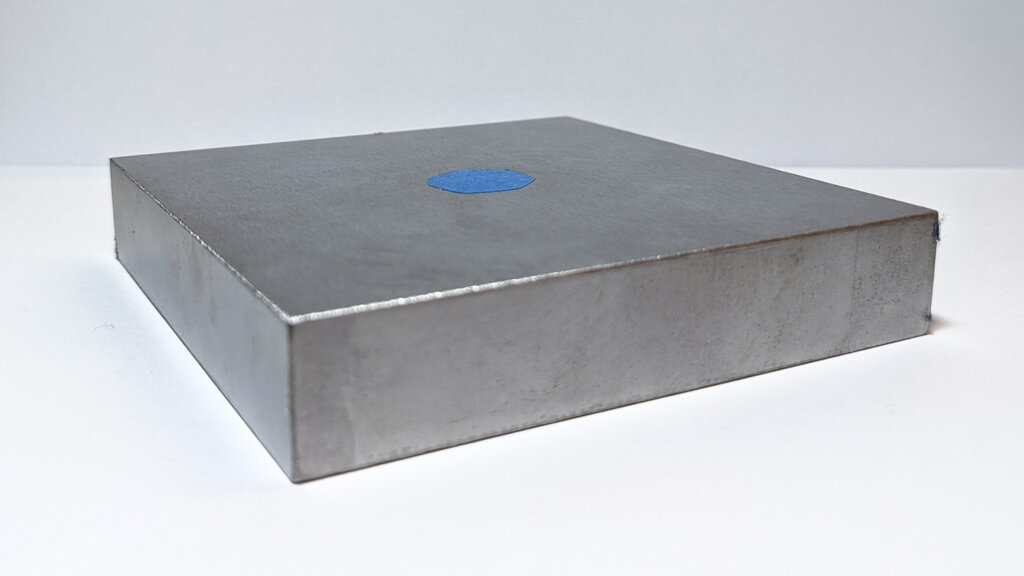}};
    \node [below=4pt of mass b, inner sep=0] (mass c) {
        \includegraphics[height=20pt]{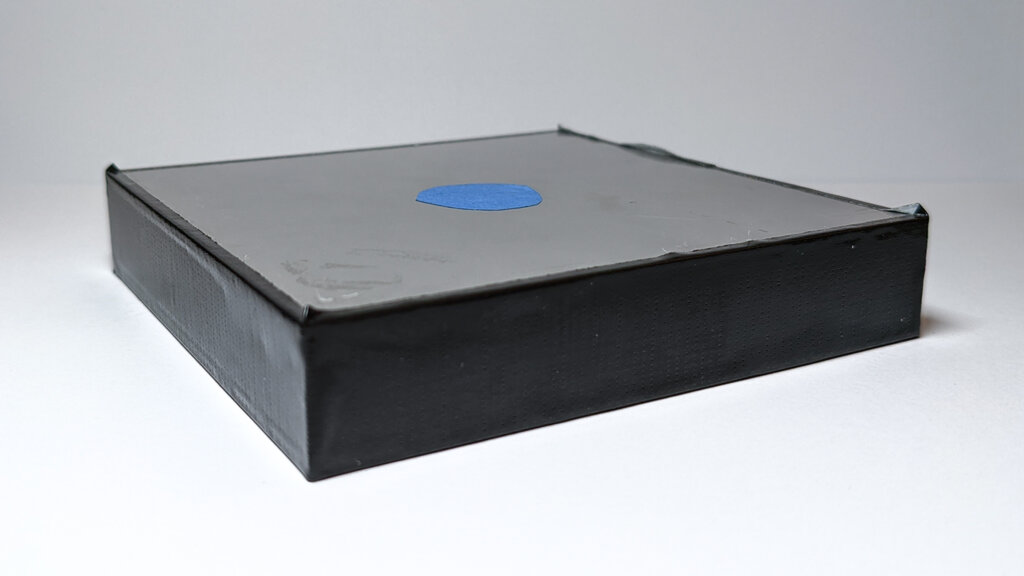}};
    \node [right=of mass a, inner sep=0] (mass d) {
        \includegraphics[height=20pt]{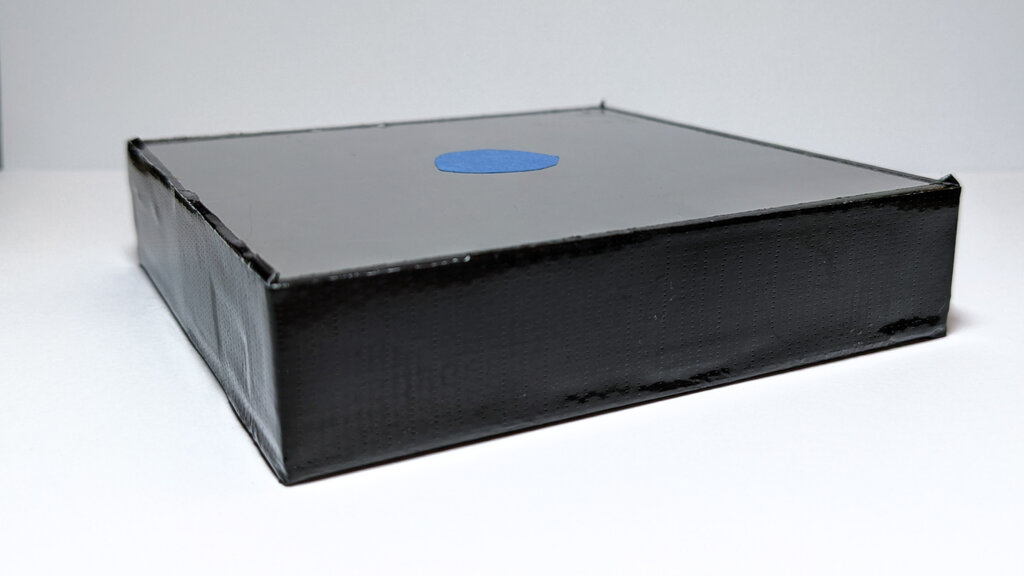}};
    \node [below=4pt of mass d, inner sep=0] (mass e) {
        \includegraphics[height=20pt]{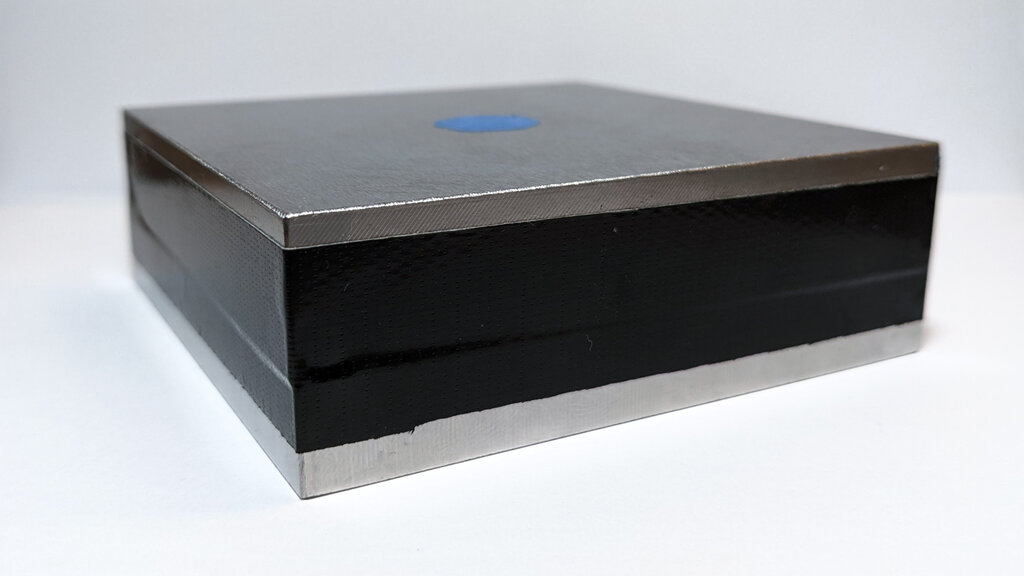}};
    \node [below=4pt of mass e, inner sep=0] (mass f) {
        \includegraphics[height=20pt]{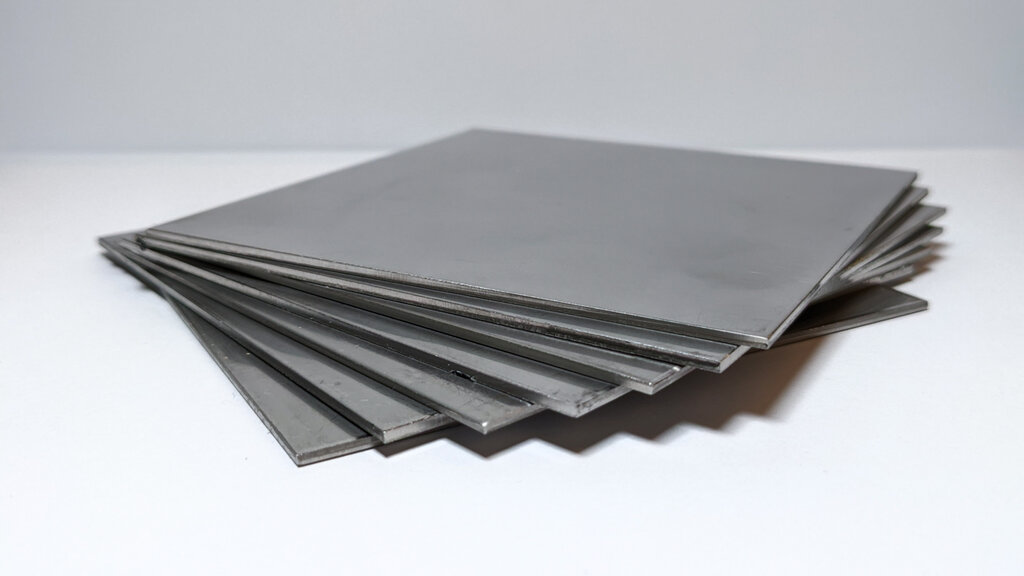}};
        
    \node [right=0pt of mass a, inner sep=1pt] { \scriptsize 1.3\,kg };
    \node [right=0pt of mass b, inner sep=1pt] { \scriptsize 1.5\,kg };
    \node [right=0pt of mass c, inner sep=1pt] { \scriptsize 1.6\,kg };
    \node [right=0pt of mass d, inner sep=1pt] { \scriptsize 1.7\,kg };
    \node [right=0pt of mass e, inner sep=1pt] { \scriptsize 2.0\,kg };
    \node [right=0pt of mass f, inner sep=1pt, align=left, font=\scriptsize] { metal \\ sheets };
    
    \end{tikzpicture}
    \caption{\textbf{Experimental setup}  In experiments, (\textbf{left}) we compute trajectories for varying horizontal distances between start and goal, and average the result.  Shorter distance require lifting faster, while longer distances can lift more gradually.  We train and test on varying masses (\textbf{right}) composed of steel blocks and stacked steel sheets.  We leave the 1.6-kg mass out of training to test generalization to unseen masses.}
    \label{fig:experimental_setup}
\end{figure}

We first perform data collection and train (Sec.~\ref{sec:training}) on four different masses, 1.343-, 1.492-, 1.741-, and 2.196-kg steel blocks (Fig.~\ref{fig:experimental_setup} right), which we round to 1.3, 1.5, 1.7, and 2\,kg hereafter. We then test \algname{} and baselines by computing trajectories between varying start and goal positions and around varying obstacles.
We automate data collection by using an overhead Intel RealSense 435i to locate the mass after suction failures.
We use a relatively heavy mass to lower the end-effector accelerations required for a lost grasp, and for safety as a lost grasp results in the released object having lower kinetic energy $E_k$, since $E_k = (1/2) m v^2$, and thus do not fly out of the workspace.

We set the automated system to collect training data for each of the four training masses, resulting in 2,367 training trajectories.  We perform an 80/20 train/test split on the trajectories, data-augment failures 30$\times$ and non-failures 4$\times$ in the training set. With the UR5 operating and generating joint data at a 125\,Hz, and the pressure sensor operating at 167\,Hz, this results in 597,010 training examples and 19,892 test examples.

With 15 trials each, we compare to baselines of J-GOMP, an optimizing motion planner that does not include inertial effects, and GOMP-FIT, an extension of J-GOMP that allows constraints on the linear acceleration at the end-effector.  We include 2 GOMP-FIT baselines in which we limit magnitude of end-effector acceleration using the unimodal model and multimodal model.

\subsection{Ablation studies}
We perform two ablation studies on \algname{}.  First, we compare history length of $h=1$ to $h=6$, to study the importance of the motion history relative to single acceleration spikes in the motion.  We also compare different values of the failure threshold $d_\mathrm{safe}$ to study the potential for making a trade-off between speed and reliability.

\input{figures/suction_grid}

\subsection{Results}

We vary the horizontal distance between start and goal configurations and compute trajectories for each baseline and variant of \algname{} (Fig.~\ref{fig:experimental_setup} left).  Each trajectory lifts over an obstacle, %
thus the shorter horizontal distance (0.8\,m) requires more vertical motion, while the longer horizontal distance (1.0\,m) results in a period of longer horizontal acceleration. 
We also vary the masses, using 3 masses that were seen during training (1.3\,kg, 1.5\,kg and 1.7\,kg) and a single mass that was not used during the data collection (1.613\,kg, rounded to 1.6\,kg).
We run each computed trajectory 5 times and report the average success rate and motion time per transported mass as the experiment's result.

In Table~\ref{tab:experiments}, we show the results of trajectories computed by the baselines and ablations.  Here we see that the J-GOMP and the GOMP-FIT unimodal baselines consistently compute motions that lead to suction failures. GOMP-FIT multimodal manages to compute safe motions, however the resulting trajectories are slow.
In contrast, \algname{} with $h=6$ and $d=0.05$ reliably achieves a 100\,\% success rate, while the baseline method suction grasps fail in nearly all cases. It is also 16\,\% to 58\,\% faster than multimodal GOMP-FIT.  The ablation of $h=1$ shows the importance of history in learning the constraint---without it, \algname{} consistently computes motions that lead to suction failures. 
In the ablation of $d_\mathrm{safe}$, we see that lower thresholds result in increased success rate, but reduced speed, while the increased $d_\mathrm{safe}$ results in faster motions and decreased success rate.
In addition, we observe that \algname{}'s performance when transporting a mass unseen during training ($m=1.6$\,kg) is comparable with its performance on the objects used during the data collection.

We also study trajectories computed with the learned constraint by varying the mass and $d_\mathrm{safe}$ and plotting the results in Fig.~\ref{fig:suction_grid}.  The plot suggests that the network is interpolating between training masses.

%% file: figures/gripper_annotated.tex
\begin{figure}[t]
    \centering
    \begin{tabular}{lr}
         \includegraphics[width=\columnwidth]{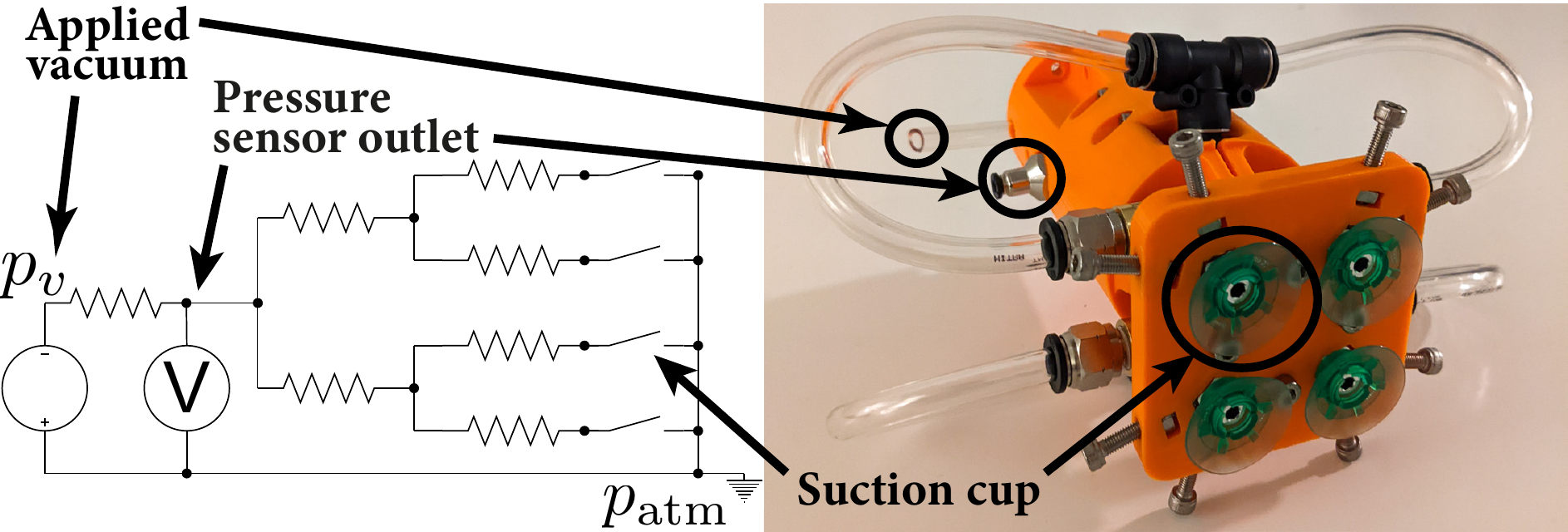}
    \end{tabular}

    \caption{\textbf{Suction gripper with four suction cups and pressure sensor outlet.}
    In the custom suction gripper (\textbf{right}), the suction is distributed to the four suction cups using branching connectors from the pneumatic tube where a vacuum/negative pressure is applied, as shown in the analogous circuit diagram (\textbf{left}). A single pressure sensor is used to measure the applied pressure. In the case of a successful suction grasp, all four switches are open, and so no current/air flows, implying the applied pressure/voltage is equal for all four suction cup regardless of differing resistor values.}
    \label{fig:gripper}
\end{figure}

%% file: table/experiments_mass.tex
\begin{table*}[t]
    \centering
    \caption{\small \textbf{Object transport success rate and motion time.}
    We compute 3 pick-and-place motions with 0.8, 0.9, and 1.0\,m horizontal separation between pick and place points, and perform each motion 5 times, for a total of 15 trials per algorithm and mass.  For algorithms, we use \textbf{J-GOMP} as a time-optimized, but not suction-constrained baseline to show a lower bound on time, \textbf{GOMP-FIT} with analytic constraints on end-effector acceleration, and \textbf{\algname{}} with varying history length $h$, and failure threshold $d_\mathrm{safe}$.  Masses 1.3, 1.5, and 1.7 were seen at train time; mass 1.6 was not.  We highlight multimodal GOMP-FIT and \algname{} $(h=6, d_\mathrm{safe}=0.50)$, and compare the relative speedup, observing that only 1 of the 60 trials for \algname{} failed, while \algname{} speeds up between 16\,\% and 58\,\%. \vspace{6pt}
    }
    \begin{tabular}{crrrrrrrr}\toprule
    \textbf{Mass} & \textbf{J-GOMP} & \multicolumn{2}{c}{\textbf{GOMP-FIT}} & \multicolumn{4}{c}{\textbf{\algname{}} $(h, d_\mathrm{safe})$} & \bf Speedup  \\
    \cmidrule(r){3-4}\cmidrule(l){5-8}
         \raisebox{6pt}[-6pt]{\bf [kg]} &  & \emph{Unimodal} & \emph{Multimodal} & $(1,0.50)$ & $(6,0.05)$ & $(6,0.50)$ & $(6,0.95)$ & \\
         \midrule
         \multicolumn{9}{c}{\textbf{Success Rate}} \\
         \midrule
         1.3\phantom{*} & 0\,\% & 0\,\% & \color{Blue} \bf 100\,\% & 0\,\% & \bf 100\,\% & \color{OliveGreen} \bf 100\,\% & 0\,\% \\
         1.5\phantom{*} & 0\,\% & 0\,\% & \color{Blue} \bf 100\,\% & 0\,\% & \bf 100\,\% & \color{OliveGreen} \bf 100\,\% & 66.7\,\%\\
         1.6* & 0\,\% & 0\,\% & \color{Blue} \bf 100\,\% & 0\,\% & \bf 100\,\% & \color{OliveGreen} 93.3\,\% & \bf 100\,\% \\
         1.7\phantom{*} & 0\,\% & 33.3\,\% & \color{Blue} \bf 100\,\% & 0\,\% & \bf 100\,\% & \color{OliveGreen} \bf 100\,\% & 33.3\,\% \\
         \midrule
         \multicolumn{9}{c}{\textbf{Motion Time [s]}} \\
         \midrule
         1.3\phantom{*} & 1.355 & 1.333 & \color{Blue}\bf 2.304 & 1.440 & 2.637 & \color{OliveGreen}\bf 1.931 & 1.728 & \bf +16\,\% \\
         1.5\phantom{*} & 1.355 & 1.387 & \color{Blue}\bf 2.827 & 1.525 & 2.763 & \color{OliveGreen}\bf 1.984 & 1.781 & \bf +30\,\% \\
         1.6*           & 1.355 & 1.472 & \color{Blue}\bf 2.827 & 1.643 & 2.795 & \color{OliveGreen}\bf 1.807 & 1.781 & \bf +36\,\% \\
         1.7\phantom{*} & 1.355 & 1.728 & \color{Blue}\bf 4.459 & 1.173 & 2.432 & \color{OliveGreen}\bf 1.856 & 1.792 & \bf +58\,\% \\
         \bottomrule
         \multicolumn{8}{l}{\footnotesize * mass unseen in training.}
    \end{tabular}
    
    \label{tab:experiments}
\end{table*}

%% file: figures/suction_grid.tex
\begin{figure}[t]
    \centering
    \includegraphics{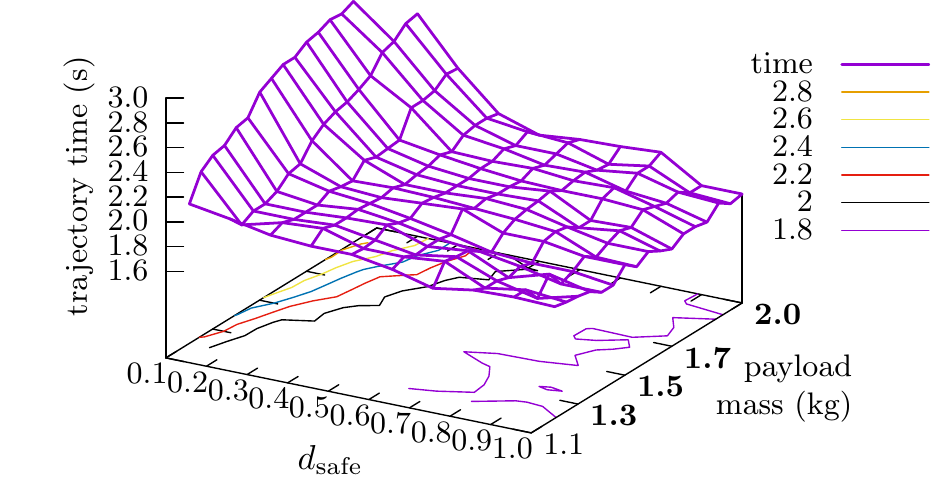}
    \caption{\textbf{Computed trajectory times for varying $d_\mathrm{safe}$ and payload mass.}  We compute trajectories for a grid of $d_\mathrm{safe}$ and masses shown on the lower axes, and plot the trajectory times on the vertical axis.  With lower $d_\mathrm{safe}$ or higher mass, the learned suction failure constraint causes the optimizer to generate slower motions.  The masses used for training (1.3, 1.5, 1.7, and 2.0\,kg) are in bold.  The plot suggests that the network generalizes to unseen masses.}
    \label{fig:suction_grid}
\end{figure}

%% file: section-6-conclusion.tex
\section{Conclusion} 
\label{sec:conclusion}

We propose \algname{}, an algorithm using learned motion constraints for fast transport of objects held in suction grasps.  By learning the constraint from real data, we avoid explicitly modeling difficult-to-model properties such as deformation of the suction cup.  We also benefit from the learned model neural-network implementation, as it facilitates automatic generation of gradients needed to linearize the constraint for the solver.  Experiments on a physical UR5 suggest that the learned constraint can allow the solver to speed up motions by up to 58\,\%. %

In future work, we will experiment with more complex environments, and include additional inputs to the learned constraint to allow it to adapt to different properties of the grasped object, for example, coefficient of friction and center of mass. We also aim to expand the analytic model to no longer be dependent on quasi-static equilibrium or rigid bodies, and therefore include spring state to the state of the trajectory optimization. While there are compelling reasons to move away from single-suction-cup grippers, they present additional suction failure modes, such as swinging and torquing out of the gripper.  Addressing these failure modes may require additional approaches, such as integration with sampling-based optimization methods.  Finally, results in modifying the failure threshold suggest that one could make a trade-off between speed and reliability, but how to beneficially make that trade-off is an open issue. 

%% file: section-7-acks.tex
\section*{Acknowledgments}
This research was performed at the AUTOLAB at UC Berkeley in
affiliation with the Berkeley AI Research (BAIR) Lab, Berkeley Deep
Drive (BDD), the Real-Time Intelligent Secure Execution (RISE) Lab, and
the CITRIS “People and Robots” (CPAR) Initiative.  We thank our colleagues for their helpful feedback and suggestions.  We thank Tae Myung Huh and Michael Danielczuk for their invaluable advice.  We thank Adam Lau for his professional photography.  We thank Adam Rashid for his help running the physical robot.  This article solely reflects the
opinions and conclusions of its authors and do not reflect the views of the sponsors or their associated entities.

%% file: main.bbl
\begin{thebibliography}{10}
\providecommand{\url}[1]{\texttt{#1}}
\providecommand{\urlprefix}{URL }
\providecommand{\doi}[1]{https://doi.org/#1}

\bibitem{acharya2020nonprehensile}
Acharya, P., Nguyen, K.D., La, H.M., Liu, D., Chen, I.M.: Nonprehensile
  manipulation: a trajectory-planning perspective. IEEE/ASME Transactions on
  Mechatronics  \textbf{26}(1),  527--538 (2020)

\bibitem{bartolini2011neuron}
Bartolini, A., Lombardi, M., Milano, M., Benini, L.: Neuron constraints to
  model complex real-world problems. In: International Conference on Principles
  and Practice of Constraint Programming. pp. 115--129. Springer (2011)

\bibitem{bernheisel2004stable}
Bernheisel, J.D., Lynch, K.M.: Stable transport of assemblies: Pushing stacked
  parts. IEEE Transactions on Automation science and Engineering
  \textbf{1}(2),  163--168 (2004)

\bibitem{berscheid2021jerk}
Berscheid, L., Kr{\"o}ger, T.: Jerk-limited real-time trajectory generation
  with arbitrary target states. arXiv preprint arXiv:2105.04830  (2021)

\bibitem{bradbury2021jax}
Bradbury, J., Frostig, R., Hawkins, P., Johnson, M.J., Leary, C., Maclaurin,
  D., Necula, G., Paszke, A., VanderPlas, J., Wanderman-Milne, S., et~al.:
  {JAX}: Autograd and {XLA}. Astrophysics Source Code Library pp. ascl--2111
  (2021)

\bibitem{deRaedt2018learning}
De~Raedt, L., Passerini, A., Teso, S.: Learning constraints from examples. In:
  Proceedings of the AAAI Conference on Artificial Intelligence. vol.~32 (2018)

\bibitem{fajemisin2021optimization}
Fajemisin, A., Maragno, D., Hertog, D.d.: Optimization with constraint
  learning: A framework and survey. arXiv preprint arXiv:2110.02121  (2021)

\bibitem{ha2021flingbot}
Ha, H., Song, S.: {FlingBot}: The unreasonable effectiveness of dynamic
  manipulation for cloth unfolding. In: Conference on Robotic Learning (CoRL)
  (2021)

\bibitem{hauser2014fast}
Hauser, K.: Fast interpolation and time-optimization with contact. The
  International Journal of Robotics Research  \textbf{33}(9),  1231--1250
  (2014)

\bibitem{huh2021multi}
Huh, T.M., Sanders, K., Danielczuk, M., Li, M., Goldberg, K., Stuart, H.S.: A
  multi-chamber smart suction cup for adaptive gripping and haptic exploration.
  arXiv preprint arXiv:2105.02345  (2021)

\bibitem{ichnowski2022gompfit}
Ichnowski, J., Avigal, Y., Liu, Y., Goldberg, K.: {GOMP-FIT}: Grasp-optimized
  motion planning for fast inertial transport. In: 2022 International
  Conference on Robotics and Automation (ICRA). IEEE (2022), \emph{(to appear)}

\bibitem{ichnowski2020djgomp}
Ichnowski, J., Avigal, Y., Satish, V., Goldberg, K.: Deep learning can
  accelerate grasp-optimized motion planning. Science Robotics  \textbf{5}(48)
  (2020)

\bibitem{ichnowski2020gomp}
Ichnowski, J., Danielczuk, M., Xu, J., Satish, V., Goldberg, K.: {GOMP}:
  Grasp-optimized motion planning for bin picking. In: 2020 International
  Conference on Robotics and Automation (ICRA). IEEE (May 2020)

\bibitem{kalakrishnan2011stomp}
Kalakrishnan, M., Chitta, S., Theodorou, E., Pastor, P., Schaal, S.: {STOMP}:
  Stochastic trajectory optimization for motion planning. In: 2011 IEEE
  international conference on robotics and automation. pp. 4569--4574. IEEE
  (2011)

\bibitem{Kavraki1996_TRA}
Kavraki, L.E., Svestka, P., Latombe, J.C., Overmars, M.: {Probabilistic
  roadmaps for path planning in high dimensional configuration spaces}. IEEE
  Trans. Robotics and Automation  \textbf{12}(4),  566--580 (1996)

\bibitem{kolluru1998}
Kolluru, R., Valavanis, K., Hebert, T.: Modeling, analysis, and performance
  evaluation of a robotic gripper system for limp material handling. IEEE
  Transactions on Systems, Man, and Cybernetics, Part B (Cybernetics)
  \textbf{28}(3),  480--486 (1998). \doi{10.1109/3477.678660}

\bibitem{kudla2018one}
Kud{\l}a, P., Pawlak, T.P.: One-class synthesis of constraints for
  mixed-integer linear programming with c4. 5 decision trees. Applied Soft
  Computing  \textbf{68},  1--12 (2018)

\bibitem{kuntz2020fast}
Kuntz, A., Bowen, C., Alterovitz, R.: Fast anytime motion planning in point
  clouds by interleaving sampling and interior point optimization. Robotics
  Research pp. 929--945 (2020)

\bibitem{LaValle2000_WAFR}
LaValle, S.M., Kuffner, J.J.: {Rapidly-exploring random trees: Progress and
  prospects}. In: Donald, B.R., Others (eds.) Algorithmic and Computational
  Robotics: New Directions, pp. 293--308. AK Peters, Natick, MA (2001)

\bibitem{lim2021prc}
Lim, V., Huang, H., Chen, L.Y., Wang, J., Ichnowski, J., Seita, D., Laskey, M.,
  Goldberg, K.: Planar robot casting with {Real2Sim2Real} self-supervised
  learning. arXiv preprint arXiv:2111.04814  (2021)

\bibitem{lombardi2017empirical}
Lombardi, M., Milano, M., Bartolini, A.: Empirical decision model learning.
  Artificial Intelligence  \textbf{244},  343--367 (2017)

\bibitem{luh1980line}
Luh, J.Y., Walker, M.W., Paul, R.P.: On-line computational scheme for
  mechanical manipulators  (1980)

\bibitem{luo2017robust}
Luo, J., Hauser, K.: Robust trajectory optimization under frictional contact
  with iterative learning. Autonomous Robots  \textbf{41}(6),  1447--1461
  (2017)

\bibitem{lynch1996dynamic}
Lynch, K.M., Mason, M.T.: Dynamic underactuated nonprehensile manipulation. In:
  Proceedings of IEEE/RSJ International Conference on Intelligent Robots and
  Systems. IROS'96. vol.~2, pp. 889--896. IEEE (1996)

\bibitem{lynch1999dynamic}
Lynch, K.M., Mason, M.T.: Dynamic nonprehensile manipulation: Controllability,
  planning, and experiments. The International Journal of Robotics Research
  \textbf{18}(1),  64--92 (1999)

\bibitem{mahler2018dex}
Mahler, J., Matl, M., Liu, X., Li, A., Gealy, D., Goldberg, K.: {Dex-Net} 3.0:
  Computing robust vacuum suction grasp targets in point clouds using a new
  analytic model and deep learning. In: 2018 IEEE International Conference on
  robotics and automation (ICRA). pp. 5620--5627. IEEE (2018)

\bibitem{maragno2021mixed}
Maragno, D., Wiberg, H., Bertsimas, D., Birbil, S.I., Hertog, D.d., Fajemisin,
  A.: Mixed-integer optimization with constraint learning. arXiv preprint
  arXiv:2111.04469  (2021)

\bibitem{mucchiani2021dynamic}
Mucchiani, C., Yim, M.: Dynamic grasping for object picking using passive
  zero-dof end-effectors. IEEE Robotics and Automation Letters  \textbf{6}(2),
  3089--3096 (2021)

\bibitem{park2012itomp}
Park, C., Pan, J., Manocha, D.: {ITOMP}: Incremental trajectory optimization
  for real-time replanning in dynamic environments. In: Twenty-Second
  International Conference on Automated Planning and Scheduling (2012)

\bibitem{pham2019critically}
Pham, H., Pham, Q.C.: Critically fast pick-and-place with suction cups. In:
  2019 International Conference on Robotics and Automation (ICRA). pp.
  3045--3051. IEEE (2019)

\bibitem{pham2013kinodynamic}
Pham, Q.C., Caron, S., Nakamura, Y.: Kinodynamic planning in the configuration
  space via admissible velocity propagation. In: Robotics: Science and Systems.
  vol.~32 (2013)

\bibitem{ratliff2009chomp}
Ratliff, N., Zucker, M., Bagnell, J.A., Srinivasa, S.: {CHOMP}: Gradient
  optimization techniques for efficient motion planning. In: 2009 IEEE
  International Conference on Robotics and Automation. pp. 489--494. IEEE
  (2009)

\bibitem{ruggiero2018nonprehensile}
Ruggiero, F., Lippiello, V., Siciliano, B.: Nonprehensile dynamic manipulation:
  A survey. IEEE Robotics and Automation Letters  \textbf{3}(3),  1711--1718
  (2018)

\bibitem{schulman2013finding}
Schulman, J., Ho, J., Lee, A.X., Awwal, I., Bradlow, H., Abbeel, P.: Finding
  locally optimal, collision-free trajectories with sequential convex
  optimization. In: Robotics: Science and Systems. pp. 1--10 (2013)

\bibitem{srinivasa2005using}
Srinivasa, S.S., Erdmann, M.A., Mason, M.T.: Using projected dynamics to plan
  dynamic contact manipulation. In: 2005 IEEE/RSJ International Conference on
  Intelligent Robots and Systems. pp. 3618--3623. IEEE (2005)

\bibitem{Stuart2015SuctionHI}
Stuart, H.S., Bagheri, M., Wang, S., Barnard, H., Sheng, A.L., Jenkins, M.,
  Cutkosky, M.R.: Suction helps in a pinch: Improving underwater manipulation
  with gentle suction flow. 2015 IEEE/RSJ International Conference on
  Intelligent Robots and Systems (IROS) pp. 2279--2284 (2015)

\bibitem{toussaint2014newton}
Toussaint, M.: Newton methods for k-order markov constrained motion problems.
  arXiv preprint arXiv:1407.0414  (2014)

\bibitem{valencia2017}
Valencia, A.J., Idrovo, R.M., Sappa, A.D., Guingla, D.P., Ochoa, D.: A 3d
  vision based approach for optimal grasp of vacuum grippers. In: 2017 IEEE
  International Workshop of Electronics, Control, Measurement, Signals and
  their Application to Mechatronics (ECMSM). pp.~1--6 (2017).
  \doi{10.1109/ECMSM.2017.7945886}

\bibitem{wang2020swingbot}
Wang, C., Wang, S., Romero, B., Veiga, F., Adelson, E.: {SwingBot: Learning
  Physical Features from In-hand Tactile Exploration for Dynamic Swing-up
  Manipulation}. In: Proc. IEEE/RSJ Int. Conf. on Intelligent Robots and
  Systems (IROS) (2020)

\bibitem{zeng2020tossingbot}
Zeng, A., Song, S., Lee, J., Rodriguez, A., Funkhouser, T.: {TossingBot}:
  Learning to throw arbitrary objects with residual physics. IEEE Transactions
  on Robotics  \textbf{36}(4),  1307--1319 (2020)

\bibitem{zhang2021rotla}
Zhang, H., Ichnowski, J., Seita, D., Wang, J., Goldberg, K.: Robots of the lost
  arc: Learning to dynamically manipulate fixed-endpoint ropes and cables. In:
  {Proc. {IEEE} Int. Conf. Robotics and Automation (ICRA)} (2021)

\end{thebibliography}
